\documentclass[11pt]{article}

\usepackage[final]{acl}

\usepackage{times}
\usepackage{latexsym}
\usepackage{amsmath}
\usepackage{amssymb}
\usepackage{amsthm}
\usepackage{booktabs}
\usepackage{subcaption}
\usepackage{algorithm}
\usepackage{algpseudocode}
\usepackage[most]{tcolorbox}
\usepackage{xcolor}
\usepackage{colortbl}
\usepackage{multirow}
\usepackage[T1]{fontenc}
\usepackage[utf8]{inputenc}

\usepackage{microtype}

\usepackage{inconsolata}

\usepackage{graphicx}
\graphicspath{{./}}

\hypersetup{hypertexnames=false}
\newtheoremstyle{compactplain}
  {2pt}
  {2pt}
  {\itshape}
  {}
  {\bfseries}
  {.}
  {0.4em}
  {}
\theoremstyle{compactplain}
\newtheorem{theorem}{Theorem}[section]
\newtheorem{proposition}[theorem]{Proposition}
\newtheorem{lemma}[theorem]{Lemma}

\theoremstyle{definition}

\newtheorem{assumption}[theorem]{Assumption}
\theoremstyle{remark}
\newtheorem{remark}[theorem]{Remark}

\title{What Drives the Zero-to-One Jump? Diagnosing and Steering Hard-Prompt Learning in RLVR}

\author{
  Rui Zhu \and
  Yumou Liu \and
  Zhenzhe Zheng\thanks{Zhenzhe Zheng is the corresponding author.} \and
  Fan Wu \\
  Shanghai Jiao Tong University \\
  Shanghai, China \\
  \texttt{\{jameszhu, liuyumou, zhengzhenzhe, wu-fan\}@sjtu.edu.cn}
}

\begin{document}
\maketitle

\begin{abstract}
Reinforcement Learning with Verifiable Rewards (RLVR) has substantially improved mathematical reasoning, yet progress on hard prompts often stalls: a prompt can remain unsolved for many training steps before abruptly transitioning to solved---a discrete ``zero-to-one'' jump whose underlying dynamics and steerability remain poorly understood. 
We investigate two questions: \textit{what drives this transition, and can it be steered?} Through event-based analysis measuring the time-to-first-positive (TTFP), we find that negative rollouts preceding successful transitions consistently retain higher early-stage generation entropy, distinguishing near-solvable failures from dead ends. This signal motivates \textbf{N}egative-\textbf{A}ugmented \textbf{P}olicy \textbf{O}ptimization (\textsc{NAPO}), a method integrated with Group Relative Policy Optimization (GRPO) that ranks negatives in unsolved groups via an early-branching driver and assigns graded advantages, injecting informative gradients precisely where standard group-relative updates vanish. Under fixed rollout budgets, \textsc{NAPO} accelerates early solution discovery and improves Pass@256 by 10.0 points on AIME24 and 6.6 points on AIME2025, demonstrating that zero-to-one transitions are diagnosable and steerable through driver-guided negative learning.
Our code is available at
\href{https://github.com/jameszhu-pixel-time/Z2O}{\texttt{https://github.com/jameszhu-pixel-time
/Z2O}}.
\end{abstract}

\begin{figure*}[t]
    \centering
    \includegraphics[width=1.0\linewidth]{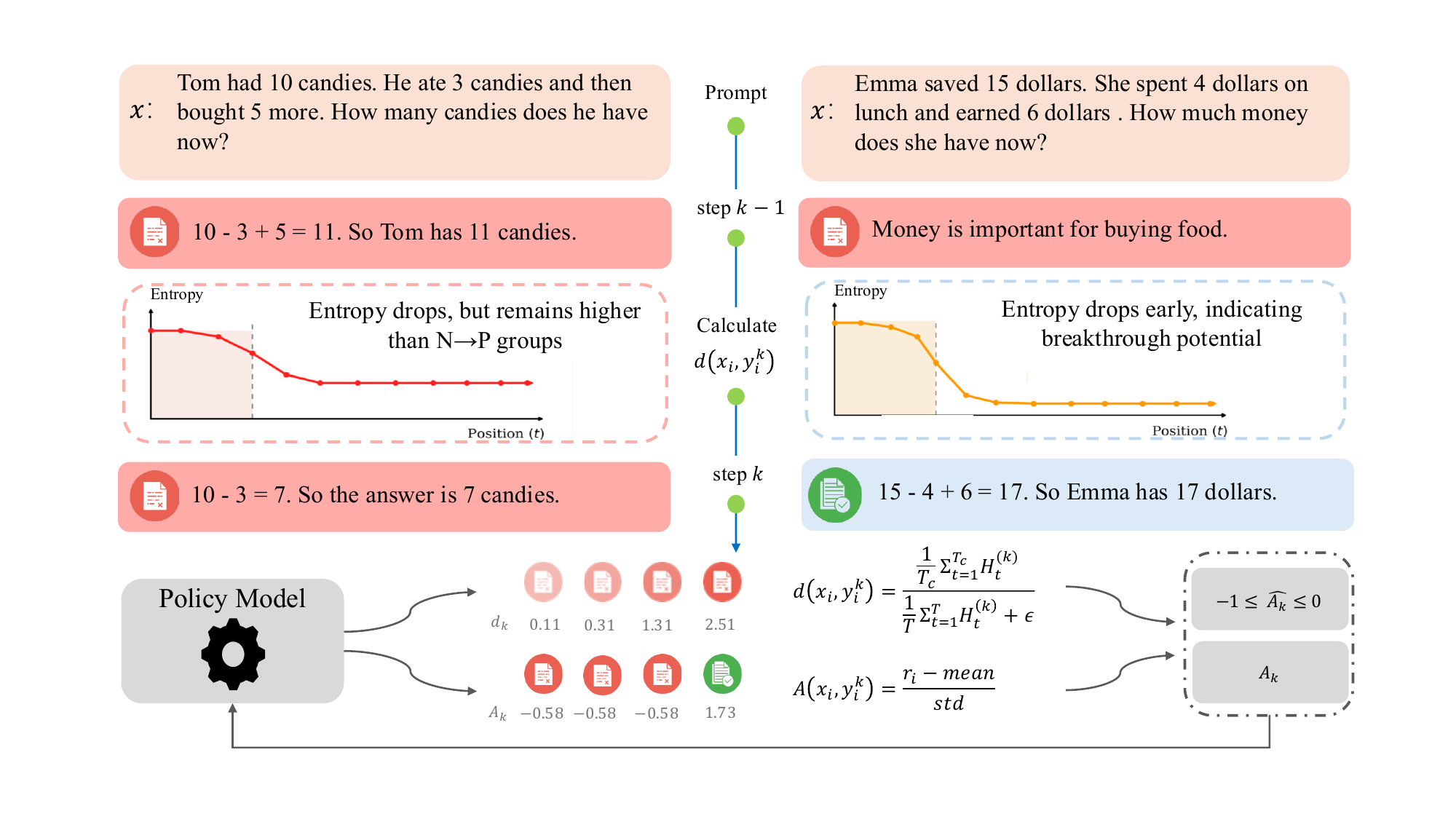}
    \caption{Overview of NAPO pipeline}
    \label{fig:enter-label}
\end{figure*}

\section{Introduction}

Reinforcement Learning with Verifiable Rewards (RLVR) \citep{lambert2025tulu3pushingfrontiers} has rapidly emerged as a dominant paradigm for post-training reasoning models, particularly in mathematical problem solving where correctness can be automatically verified \citep{yang2025qwen3technicalreport,guo2025deepseekr1}. 
At each training step, the policy samples multiple rollouts per prompt and receives binary rewards indicating whether each final answer is correct.
When some rollouts in a group succeed, the resulting contrast between positive and negative samples provides a clear gradient that drives steady improvement. 
However, on sufficiently hard prompts the on-policy pass rate is so low that entire rollout groups contain only incorrect samples for many consecutive steps.
In group-relative methods such as GRPO~\citep{deepseek-math,guo2025deepseekr1}, this eliminates reward contrast and yields near-zero advantages on precisely the prompts that most need exploration. 
As a result, hard prompts often remain stagnant for many steps and then transition abruptly from unsolved to solved — a discrete ``zero-to-one'' jump rather than smooth refinement. The dynamics that drive this transition, and whether they can be steered, remain largely unexplored.

Answering these questions, however, requires confronting two intertwined obstacles that current RLVR pipelines leave unresolved. The first obstacle is diagnosing zero-to-one events under binary supervision. Standard RLVR diagnostics offer little purchase on what happens inside a hard prompt before it is first solved: pass@k \citep{chen2021evaluatinglargelanguagemodels} aggregates end-of-training capability and cannot localize the moment a previously unsolved prompt becomes solvable, while group-level rewards remain identically zero across long stretches of training and therefore carry no information about which unsolved prompts are close to a breakthrough. What is missing is (i) an event-based view of training that pinpoints when transitions occur and (ii) an internal, rollout-level signal that distinguishes near-solvable failures from true dead ends using only the information already available at sampling time.

The second obstacle is turning heterogeneous negatives into informative gradient on fully-negative groups. A majority of RLVR methods build on grouped objectives such as GRPO and its variants \citep{deepseek-math,yu2025dapoopensourcellmreinforcement,zheng2025groupsequencepolicyoptimization,wu2026quantileadvantageestimationstabilizing}, where all-negative groups produce vanishing advantages and provide little learning signal \citep{ma2026tspobreakingdoublehomogenization,yu2025dapoopensourcellmreinforcement}. Recent variants restore a gradient through weighted negative learning, virtual rewards, or entropy-aware updates \citep{NEURIPS2025_b8026fe0,nan2025ngrponegativeenhancedgrouprelative,le2026promptleftbehindexploiting}, but they apply roughly uniform pressure across negatives.
The open problem is therefore how to assign graded credit among them under only binary rewards, in a way that injects gradient into unsolved groups without destabilizing the rest of training.

We analyze token-level generation entropy during RLVR training and find that negative rollouts from prompts that later undergo a Negative-to-Positive (N$\rightarrow$P) transition exhibit front-loaded uncertainty, with early-response entropy patterns that distinguish them from rollouts on prompts that remain negative (N$\rightarrow$N). Based on this signal, we propose NAPO (Negative-Augmented Policy Optimization), a GRPO-integrated method that ranks negative samples by front-loaded entropy and injects informative gradients into unsolved groups.

Our contributions are threefold:
\begin{itemize}
  \item We study zero-to-one learning dynamics in RLVR through event-based metrics, especially \textbf{N$\rightarrow$P} transitions and \textbf{TTFP}, to characterize how hard prompts first become solvable.

  \item We identify a token-entropy pattern associated with N$\rightarrow$P transitions and propose early-branching drivers that rank negative rollouts by their latent learning potential.

  \item We introduce NAPO, a GRPO-integrated algorithm that assigns graded advantages to negative samples according to their driver-based ranking while maintaining stable updates via in-group normalization, improving token-efficient exploration on hard prompts as reflected by better TTFP and higher pass@k.
\end{itemize}

\section{Preliminaries}\label{sec:preliminaries}\label{sec:dist-shift}

\paragraph{Related Work.}
Reinforcement learning (RL) \citep{sutton1998reinforcement} has been shown to improve LLM reasoning via both exploration and exploitation~\citep{auer2002finite,cheng2025reasoningexplorationentropyperspective}. Approaches such as RLHF~\citep{ouyang2022traininglanguagemodelsfollow}, Test-Time RL~\citep{zuo2025ttrltesttimereinforcementlearning}, and RLVR~\citep{guo2025deepseekr1} train policies using verifiable rewards or reward models, often with policy-gradient methods like PPO, GRPO, and REINFORCE~\citep{schulman2017ppo,deepseek-math,williams1992simple}. These methods can stall on very hard prompts~\citep{ma2026tspobreakingdoublehomogenization}. Recent work addresses this by reallocating rollouts toward learnable prompts and by extracting informative signals from negative samples, including token-level entropy to encourage exploration and reduce overfitting~\citep{NEURIPS2025_b8026fe0,le2026promptleftbehindexploiting,fang2026allocatelearndynamicrollout}. We further introduce the \textbf{negative-to-positive transition} to predict and leverage latent prompt learnability to overcome this challenge.

\paragraph{Group-Relative Policy Optimization for RLVR .}
Following recent RLVR practice for reasoning models, we sample a group of $G$ rollouts $\{y_i\}_{i=1}^G$ for each prompt and optimize the policy with Group Relative Policy Optimization (GRPO) \citep{guo2025deepseekr1,deepseek-math}.
Let $\pi_{\theta_{\text{old}}}$ be the snapshot policy and define the token-level likelihood ratio
\(
\rho_{i,t}(\theta) = \frac{\pi_\theta(a_{i,t}\mid s_{i,t})}{\pi_{\theta_{\text{old}}}(a_{i,t}\mid s_{i,t})}.
\)
GRPO then applies the clipped surrogate objective 
\begin{equation}
\label{eq:grpo}
\begin{aligned}
\max_\theta\;&\mathbb{E}_{x}\mathbb{E}_{\ y_i \sim \pi_{\theta_{\mathrm{old}}}}
\Bigg[\sum_{i=1}^{G}\sum_t \ell_{i,t}
-\beta\,\mathrm{KL}(\pi_\theta\,\|\,\pi_{\mathrm{ref}})\Bigg],\\
\ell_{i,t}&=\min\Big(\rho_{i,t}(\theta)\hat{A}_{i,t},\\
&\quad \mathrm{clip}(\rho_{i,t}(\theta),1-\epsilon,1+\epsilon)\; \hat{A}_{i,t}\Big),
\end{aligned}
\end{equation}
where $\hat{A}_{i,t}$ is the group-relative advantage, $\pi_{\mathrm{ref}}$ is a reference policy, and $\beta$ controls KL regularization. For brevity, we use $\pi_\theta$ and $\pi_{\mathrm{ref}}$ to denote the distributions conditioned on a given prompt $x$, i.e., $\pi_\theta(\cdot \mid x)$ and $\pi_{\mathrm{ref}}(\cdot \mid x)$, respectively.

\paragraph{PSR/NSR decomposition.}
For verifiable-reward post-training, each rollout receives a binary outcome. We view the learning signal through the positive-sample reward (PSR) and negative-sample reward (NSR) decomposition \citep{NEURIPS2025_b8026fe0}. Let $r(x,y)\in\{-1,+1\}$ be the verifiable reward for completion $y$ on prompt $x$. A simplified RLVR objective is
\begin{equation}
\label{eq:rlvr_sign_split}
\mathcal{L}_{\mathrm{RLVR}}(\theta)
\;:=\;
-\mathbb{E}_{x\sim\mathcal{D}}\,
\mathbb{E}_{y\sim\pi_\theta(\cdot\mid x)}\!\left[r(x,y)\right].
\end{equation}
Using $r = \mathbb{I}[r=+1]-\mathbb{I}[r=-1]$, we can write
\begin{equation}
\label{eq:psr_nsr_split}
\begin{aligned}
\mathcal{L}_{\mathrm{RLVR}}(\theta)
=&\;
\underbrace{
-\mathbb{E}_{x\sim\mathcal{D}}\,
\mathbb{E}_{y\sim\pi_\theta(\cdot\mid x)}
\!\left[\mathbb{I}[r(x,y)=+1]\right]
}_{\mathcal{L}_{\mathrm{PSR}}(\theta)}
\\
&\quad+
\underbrace{
\mathbb{E}_{x\sim\mathcal{D}}\,
\mathbb{E}_{y\sim\pi_\theta(\cdot\mid x)}
\!\left[\mathbb{I}[r(x,y)=-1]\right]
}_{\mathcal{L}_{\mathrm{NSR}}(\theta)}.
\end{aligned}
\end{equation}
PSR reinforces verified rollouts, whereas NSR suppresses failed ones. 
Both sides can provide useful superivision on reasoning tasks \citep{NEURIPS2025_b8026fe0}.

\paragraph{Token-level uncertainty and rollout-level drivers.}
To analyze how N$\to$P happens, we analyze the patterns of negative rollouts before N$\to$P happens.
We measure token-level uncertainty with the entropy of the next-token distribution \citep{wang2025beyond8020,yang2025letitcalm}.
For each rollout, let $p_t^{(k)}$ denote the top-$k$ truncated next-token distribution at position $t$ after renormalization, with invalid and padding positions masked out. We define the token-level entropy sequence
\begin{equation}
H_t^{(k)}=-\sum_{v} p_t^{(k)}(v)\log p_t^{(k)}(v),
\end{equation}
and its exponential form
\begin{equation}\label{eq:rho}
\rho_t^{(k)}=\exp\!\big(H_t^{(k)}\big),
\end{equation}
which we use as a token-level perplexity / effective branching statistic. Our goal is not to measure absolute early-stage uncertainty in isolation, but to summarize the \emph{relative prominence} of early uncertainty within each rollout.
\footnote{In off-policy settings, rollouts may come from another policy $\pi_\phi$, but unless stated otherwise we still compute $H_t$ under the training policy $\pi_\theta$. $k=20$ unless otherwise stated.}

\paragraph{N$\to$P Transition.}
To localize progress on hard prompts, let $q_t(x)$ indicate whether evaluation at training step $t$ contains at least one positive rollout for prompt $x$. We define the time-to-first-positive (TTFP) as
\begin{equation*}
\mathrm{TTFP}(x) \;=\; \min \{ t \,:\, q_t(x)=1 \},
\end{equation*}
with $\mathrm{TTFP}(x)=\infty$ if no positive appears within the training horizon. We define an N$\rightarrow$P transition as
\begin{equation*}
q_{t-1}(x)=0 \ \wedge\ q_t(x)=1.
\end{equation*}
This implies that the model solves a problem on which all previous attempts failed.

\section{Signals of Near-Solvable Failures}
\label{sec:early-branch}

This section answers the question of what drives the N$\to$P transition.

\subsection{Experimental Setup}
\label{sec:setup}

\begin{table*}[tbp]
\centering
\small
\setlength{\tabcolsep}{5.5pt}
\resizebox{\textwidth}{!}{%
\begin{tabular}{lcc cc cc ccc}
\toprule
& \multicolumn{2}{c}{AMC23}
& \multicolumn{2}{c}{AIME2025}
& \multicolumn{2}{c}{Math-Beyond}
& \multicolumn{3}{c}{Training Set} \\
\cmidrule(lr){2-3}
\cmidrule(lr){4-5}
\cmidrule(lr){6-7}
\cmidrule(lr){8-10}
Replay
& N$\rightarrow$P & P$\rightarrow$N
& N$\rightarrow$P & P$\rightarrow$N
& N$\rightarrow$P & P$\rightarrow$N
& N$\rightarrow$P & P$\rightarrow$N & TTFP \\
\midrule
\textit{n1} 
& \textbf{2}/3 & 0.33 
& \textbf{2.67}/20 & 2 
& \textbf{12}/134 & 5.33 
& \textbf{120} & 116 & \textbf{1.213} \\
\textit{n2} 
& 1.67/3 & 2.67 
& 2/20 & 2.33 
& 8.33/134 & 5.33 
& 117.67 & 102.67 & 1.225 \\
\bottomrule
\end{tabular}%
}
\caption{\textbf{Transition statistics at $t^*=1$ for the \textit{n1}--\textit{n2} replay pair.}
A prompt is labeled as $P$ if at least one rollout is correct under the fixed rollout budget, and as $N$ otherwise.
In the validation columns, $a/b$ reports $a$ N$\rightarrow$P transitions among $b$ prompts that are negative at the base checkpoint.
Results are averaged over three runs, and TTFP is computed over the first three steps. Best value is shown in bold.}
\label{tab:transition-stats}
\end{table*}
We examine two axes of variation in this section: (i) \emph{within-batch}, comparing rollouts inside all-negative groups, and (ii) \emph{between-replay}, comparing two training continuations that share a base state but differ in sampling stochasticity. 
\paragraph{Fixed-batch protocol.}
RLVR on hard prompts exhibits strong temporal correlations across training steps \citep{li2025limrrlscaling,zheng2025actpaysefficientreinforcement}. 
To expose this structure directly, we train Qwen2.5-Math-7B on a fixed set of $1024$ DAPO-17k prompts for $10$ consecutive gradient steps, sampling $G = 8$ rollouts per prompt per step. 
Holding the prompt batch fixed across steps lets us align rollout-internal uncertainty patterns with later N$\to$P events using the metrics of \S\ref{sec:preliminaries}. 
Unless otherwise noted, we use temperature $T = 1.0$, maximum context length $4096$, learning rate $10^{-6}$, and zero entropy/KL penalties.\footnote{Full hyperparameters appear in Appendix~\ref{D.1.} and Table~\ref{tab:hyperparams_grpo}.}

\paragraph{Critical-step replay.}
Across the $10$-step window, we identify the step $t^{*}$ at which N$\to$P events peak and treat it as the \emph{critical checkpoint}.
Starting from this shared state, we replay training with the same base policy, optimizer state, data order, and hyperparameters, varying only rollout stochasticity. 
This yields two representative continuations, denoted $n_1$ and $n_2$. 
We do not label either replay as ``good'' or ``bad'' at this point.

\paragraph{Outcome measures.}
\emph{(i) Prompt-level transition label.} Among prompts that are negative at the base checkpoint $t^{*}$, we call a prompt \textbf{N$\to$P} if at least one later checkpoint within the replay window produces a positive rollout for it, and \textbf{N$\to$N} otherwise. 
Each negative rollout inherits its prompt's label. 
At training time the binary reward gives both groups identical labels; the partition into N$\to$P and N$\to$N is only observable \emph{after} the replay window, which is precisely what makes the within-batch driver signal of \S\ref{sec:within-batch} non-trivial.

\emph{(ii) Replay-level transition counts.} 
For each replay we report the number of N$\to$P and P$\to$N transitions under $64$ rollouts per prompt, together with the mean TTFP over prompts within the replay window. Smaller first-positive indices indicate earlier solution discovery.

\emph{(iii) Validation sets.} We evaluate replay checkpoints on AMC23, AIME~2025, and Math-Beyond \citep{mayilvahanan2025mathbeyondbenchmarkrlexpand} under the same $64$-rollout budget.

\subsection{Findings and Implications}
\label{sec:findings}

We first define a rollout-time driver that compresses the entropy--position pattern into a single scalar (\S\ref{sec:driver-def}), then validate it at each level in turn (\S\ref{sec:within-batch} and \S\ref{sec:between-replay}).

\begin{figure*}[h]
    \centering
    \begin{subfigure}{0.48\linewidth}
        \centering
        \includegraphics[width=0.9\linewidth]{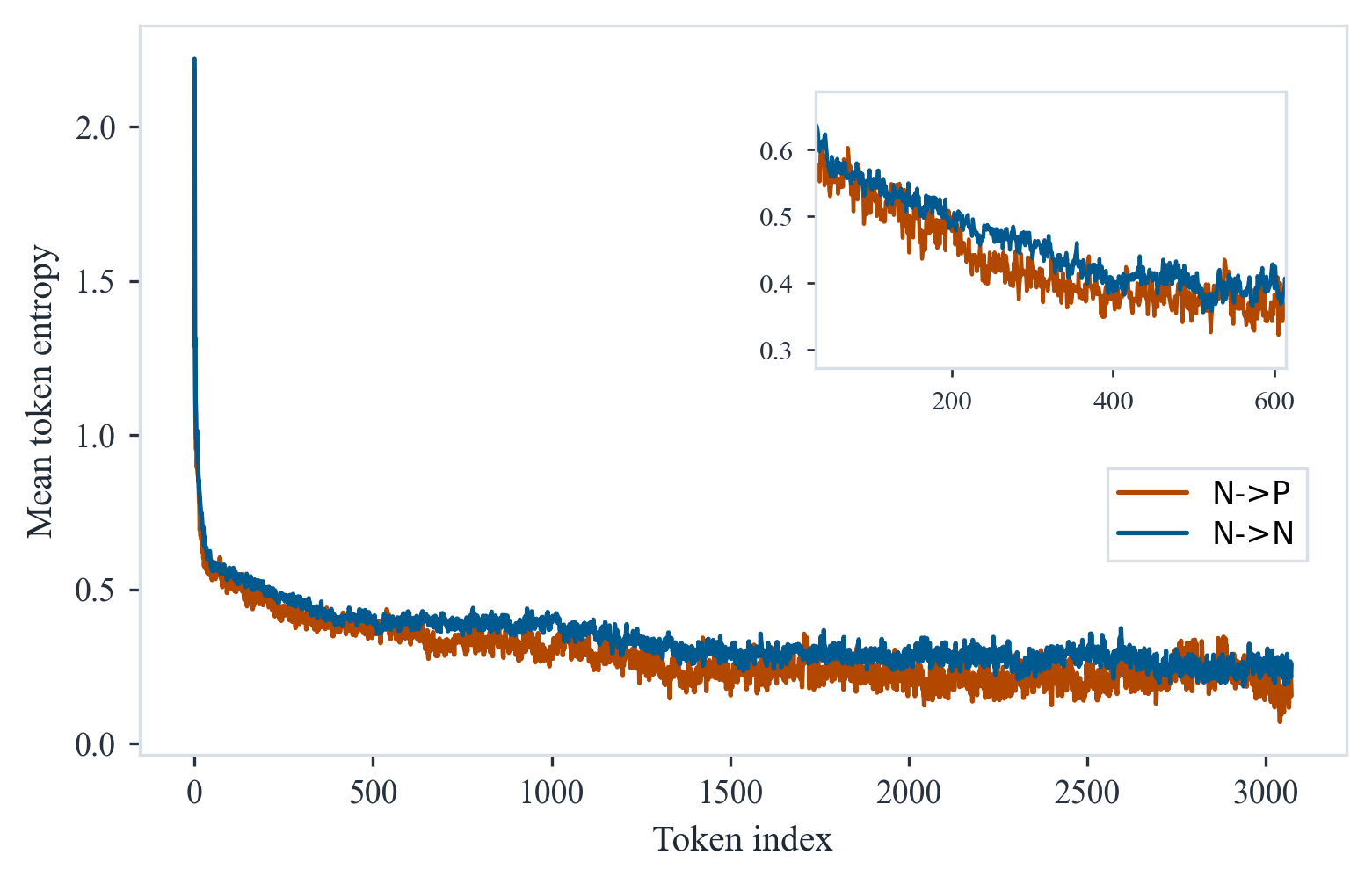}
        \caption{Entropy-position curves for the bad replay \textit{n2}.}
        \label{fig:token_pos}
    \end{subfigure}
    \hfill
    \begin{subfigure}{0.48\linewidth}
        \centering
        \includegraphics[width=0.9\linewidth]{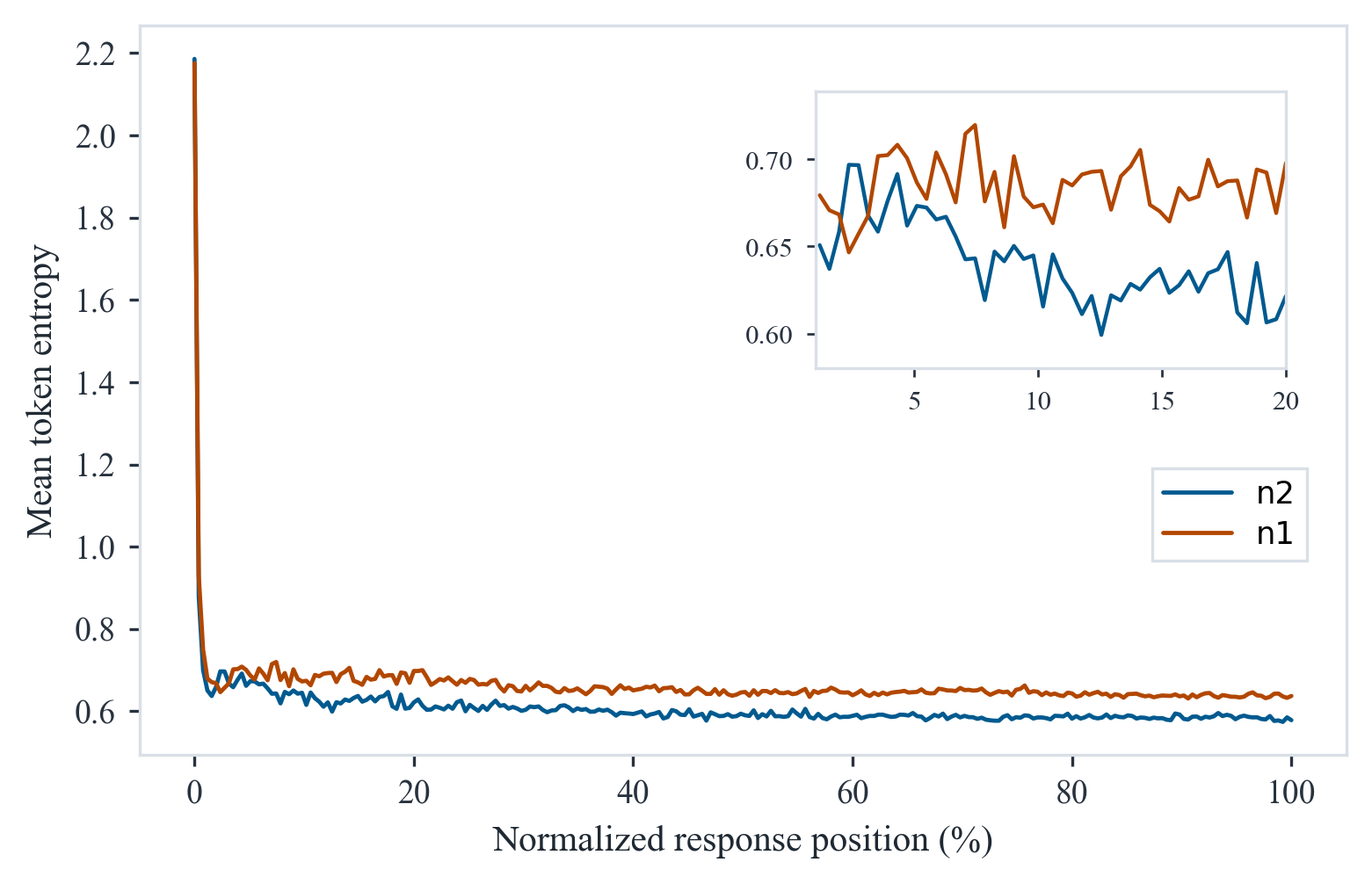}
        \caption{Comparison of the N$\rightarrow$P group between \textit{n1} and \textit{n2}.}
        \label{fig:ent_position}
    \end{subfigure}
    \caption{\textbf{Entropy-position plots.} The main separation is localized in the early part of generation. Prompts that later undergo N$\rightarrow$P transitions preserve more early-stage uncertainty than prompts that remain N$\rightarrow$N, and this pattern is sharper in the good replay \textit{n1} than in the bad replay \textit{n2}.}
    \label{fig:entropy_token_side_by_side}
\end{figure*}

\subsubsection{An early-branching driver}
\label{sec:driver-def}

\paragraph{Why a ratio, not an absolute level.}
A naive metric would be the average entropy of a rollout, but the contrast we care about is not ``more entropy'' in absolute terms; it is a difference in the \emph{timing} of uncertainty---how much of a rollout's total uncertainty is allocated to its opening tokens \cite{fang2026allocatelearndynamicrollout}. 

\paragraph{Definition.}
Let $T_{\alpha} = \lceil \alpha\,T \rceil$ denote the number of tokens in the
first $\alpha$ fraction of a response of length $T$. The
\emph{early mean-perplexity ratio} is
\begin{equation}
d_{\text{early-ppl}} \;=\;
\frac{\tfrac{1}{T_{\alpha}}\sum_{t=1}^{T_{\alpha}} \rho_t^{(k)}}
     {\tfrac{1}{T}\sum_{t=1}^{T} \rho_t^{(k)}},
\label{eq:early_ratio}
\end{equation}
where $\rho_t^{(k)} = \exp\!\bigl(H_t^{(k)}\bigr)$ is the exponential of the top-$k$ truncated entropy defined in Eq.~\eqref{eq:rho}. Values of $d_{\text{early-ppl}} > 1$ indicate that uncertainty is concentrated early in the response\footnote{
Since entropy is measured on a logarithmic scale, $\exp(H_t^{(k)})$ can be interpreted as a top-$k$ truncated local perplexity, i.e., the effective branching factor of plausible next-token continuations. This is related to the branching-factor view in \citep{yang2025llmprobabilityconcentrationalignment}.
}. 

\paragraph{Choice of the early fraction.}We introduce the early-fraction parameter $\alpha$ and compare $0.1$, $0.2$, and $0.5$ on the $n_1$--$n_2$ replay pair (Fig.~\ref{fig:ng_training})\footnote{Details are deferred to \S \ref{sec:Z2O}.}. Among these candidates, $0.2$
yields the clearest and most stable separation between N$\to$P and N$\to$N rollouts while remaining robust across both replays, and we use it throughout the remainder of the paper.

\begin{figure*}[t]
  \centering
  \begin{subfigure}[t]{0.48\linewidth}
    \centering
    \includegraphics[width=0.9\linewidth]{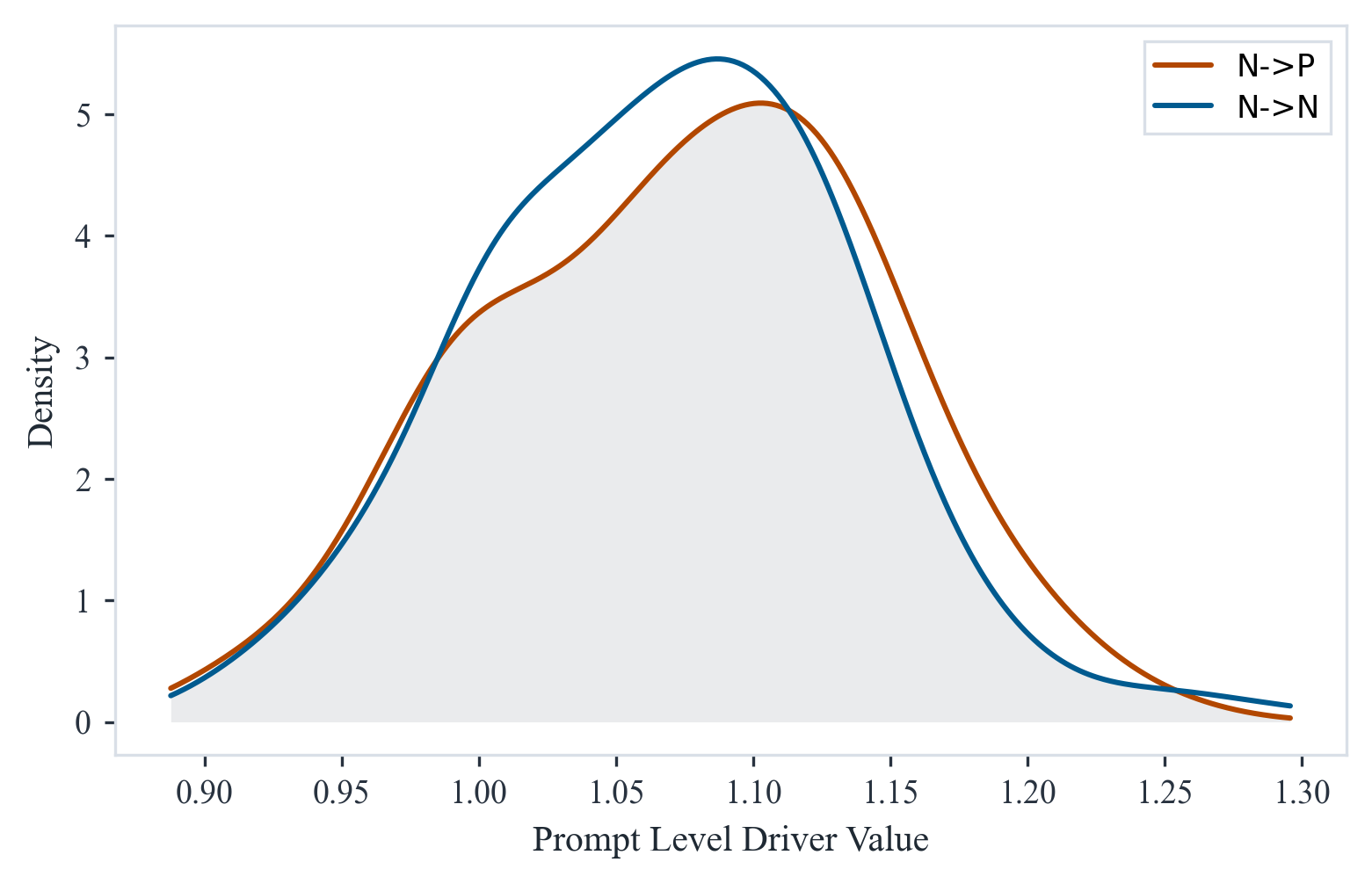}
    \caption{Driver distribution for the good replay \textit{n1}. See Appendix~\ref{app:driver_stats} for statistics' demonstration}
    \label{fig:early_ppl_inbatch_n1}
  \end{subfigure}\hfill
  \begin{subfigure}[t]{0.48\linewidth}
    \centering
    \includegraphics[width=0.9\linewidth]{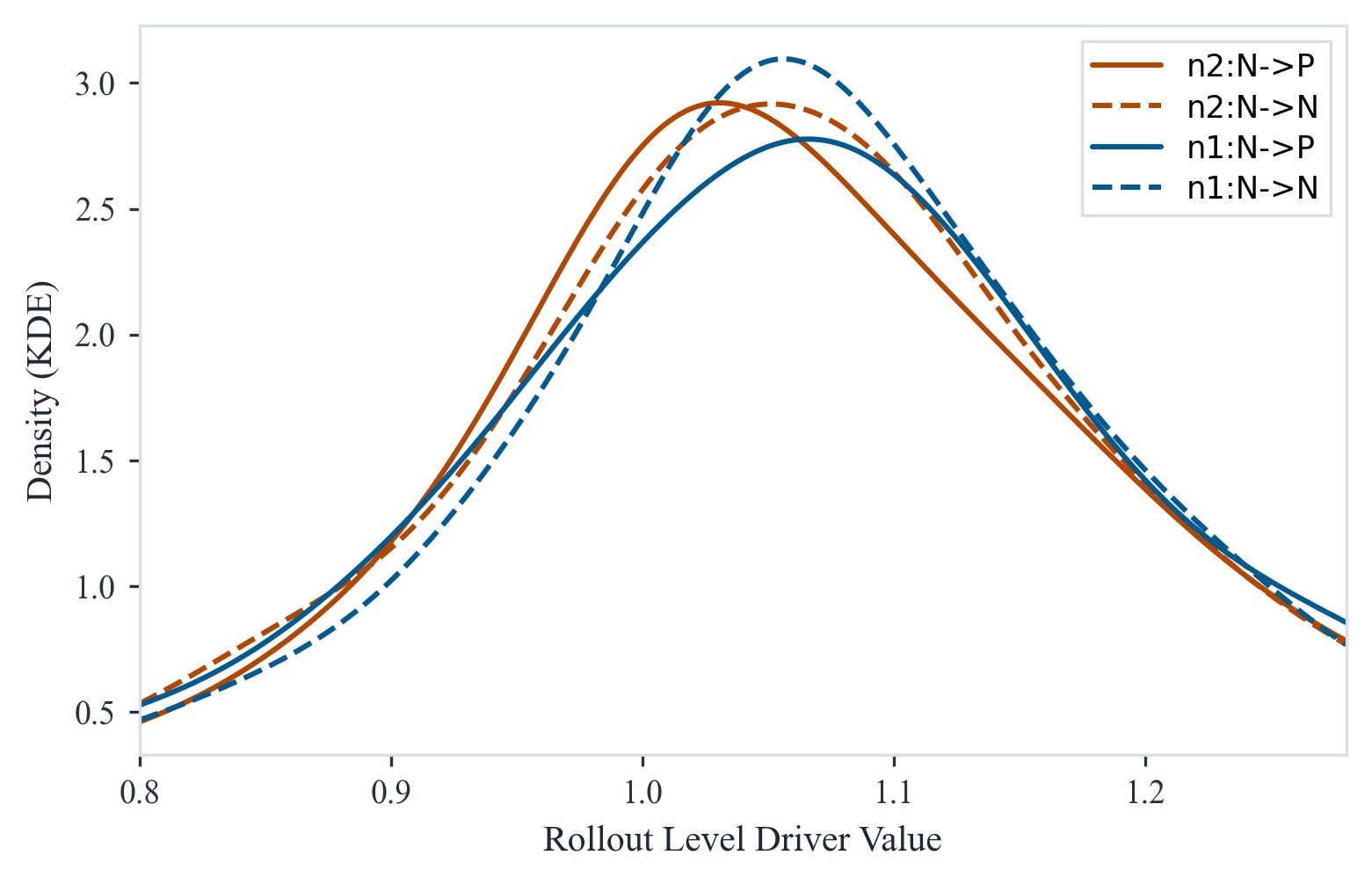}
    \caption{Driver distribution for the bad replay \textit{n2} and good replay \textit{n1}.}
    \label{fig:early_ppl_inbatch_n1n2}
  \end{subfigure}
  \caption{\textbf{Within-batch separation under the practical driver $d_{\text{early-ppl}}$.} Rollouts associated with prompts that later undergo N$\rightarrow$P transitions are systematically right-shifted relative to rollouts from prompts that remain N$\rightarrow$N. The separation is present in both replays, but is cleaner and more pronounced in the good replay \textit{n1}.}
  \label{fig:early_ppl_inbatch}
\end{figure*}

\subsubsection{Within-batch evidence: negatives are not homogeneous}
\label{sec:within-batch}

\paragraph{Entropy--position pattern.}
Figure~\ref{fig:token_pos} plots mean token entropy as a function of position for negative rollouts in replay $n_2$. The N$\to$P group maintains noticeably higher uncertainty at the beginning of the response, whereas the N$\to$N group collapses earlier to lower-entropy continuations. The two groups converge later in the response, confirming that the discriminative signal is localized in the early stage.

\paragraph{Driver separation.}
Figure~\ref{fig:early_ppl_inbatch_n1} shows the distribution of $d_{\text{early-ppl}}$ for the two groups in replay $n_1$. Negatives associated with future N$\to$P prompts are systematically shifted toward larger driver values than negatives associated with N$\to$N prompts, mirroring the qualitative pattern of Fig.~\ref{fig:token_pos} under the single scalar of Eq.~\eqref{eq:early_ratio}.

\paragraph{Claim A (within-batch).}
The driver provides a rollout-level ordering signal that separates near-solvable failures from dead ends inside an otherwise indistinguishable negative pool.

\subsubsection{Between-replay evidence: the driver tracks aggregate transition tendency}
\label{sec:between-replay}

\paragraph{Entropy--position pattern.}
Figure~\ref{fig:ent_position} restricts attention to N$\to$P negatives and compares $n_1$ against $n_2$. The good replay $n_1$ preserves more early-stage uncertainty than the bad replay $n_2$, sharpening the front-loaded pattern of Claim~A but now across replays rather than within one. This is consistent with the relevant contrast being a matter of timing rather than magnitude: the same shape of distribution is shifted earlier in $n_1$ than in $n_2$.
\paragraph{Driver distribution.}
Figure~\ref{fig:early_ppl_inbatch_n1n2} overlays all four (replay $\times$ label) groups. $n_1$ allocates more probability mass to high-driver rollouts than $n_2$ does in both groups. \footnote{Batch-level summary statistics quantifying this right-tail dominance are reported in Appendix~\ref{app:driver_stats}.}
\paragraph{Realized transitions.}
Table~\ref{tab:transition-stats} reports the downstream outcome. Starting from an identical base state, the good replay $n_1$ realizes more N$\to$P transitions on the training set and attains a smaller mean first-positive index than $n_2$. 
The ranking of realized transitions matches the ranking of right-tail driver mass, indicating that the batch-level driver shift is not merely a cosmetic difference but is related to \emph{which} replay will subsequently break more prompts out of the all-negative regime. 
This is consistent with recent RLVR observations that low-success prompts exhibit structured temporal dynamics rather than noise-like fluctuations throughout training \citep{zheng2025actpaysefficientreinforcement}.
\paragraph{Claim B (between-replay).}
The driver also functions as a batch-level diagnostic: batches whose
driver distribution has a heavier right-tail mass go on to realize
more N$\to$P events.
\begin{tcolorbox}[colback=gray!5,colframe=black!70,title=\textbf{Takeaway},fonttitle=\bfseries]
Prompts approaching the N$\to$P boundary exhibit a distinctive form of front-loaded uncertainty in their negative rollouts. The early-branching driver $d_{\text{early-ppl}}$ captures this pattern at two levels of aggregation: it separates N$\to$P from N$\to$N negatives \emph{within} a batch (Claim~A), and it separates higher-yield from lower-yield batches \emph{across} replays (Claim~B). 
\end{tcolorbox}

\begin{figure*}[htbp]
    \centering
    \begin{minipage}{0.47\textwidth}
        \centering
        \includegraphics[width=\linewidth]{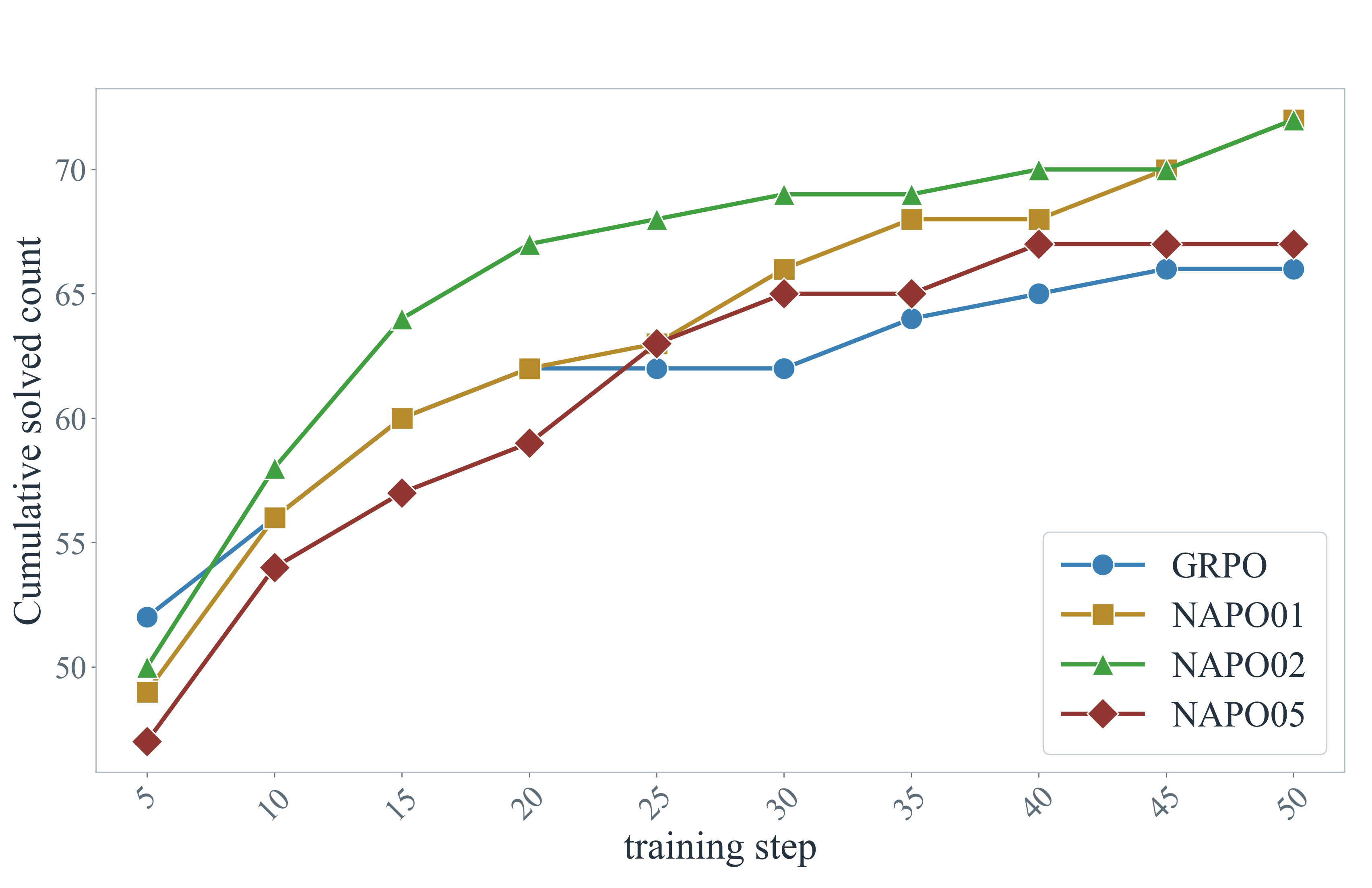}
        \caption{Cumulative solved prompts on validation sets. NAPO achieves faster early progress, solving more prompts within the first 50 steps, while maintaining comparable performance at later stages.}
        \label{fig:ttfp_nvsg}
    \end{minipage}
    \hfill
    \begin{minipage}{0.47\textwidth}
        \centering
        \includegraphics[width=\linewidth]{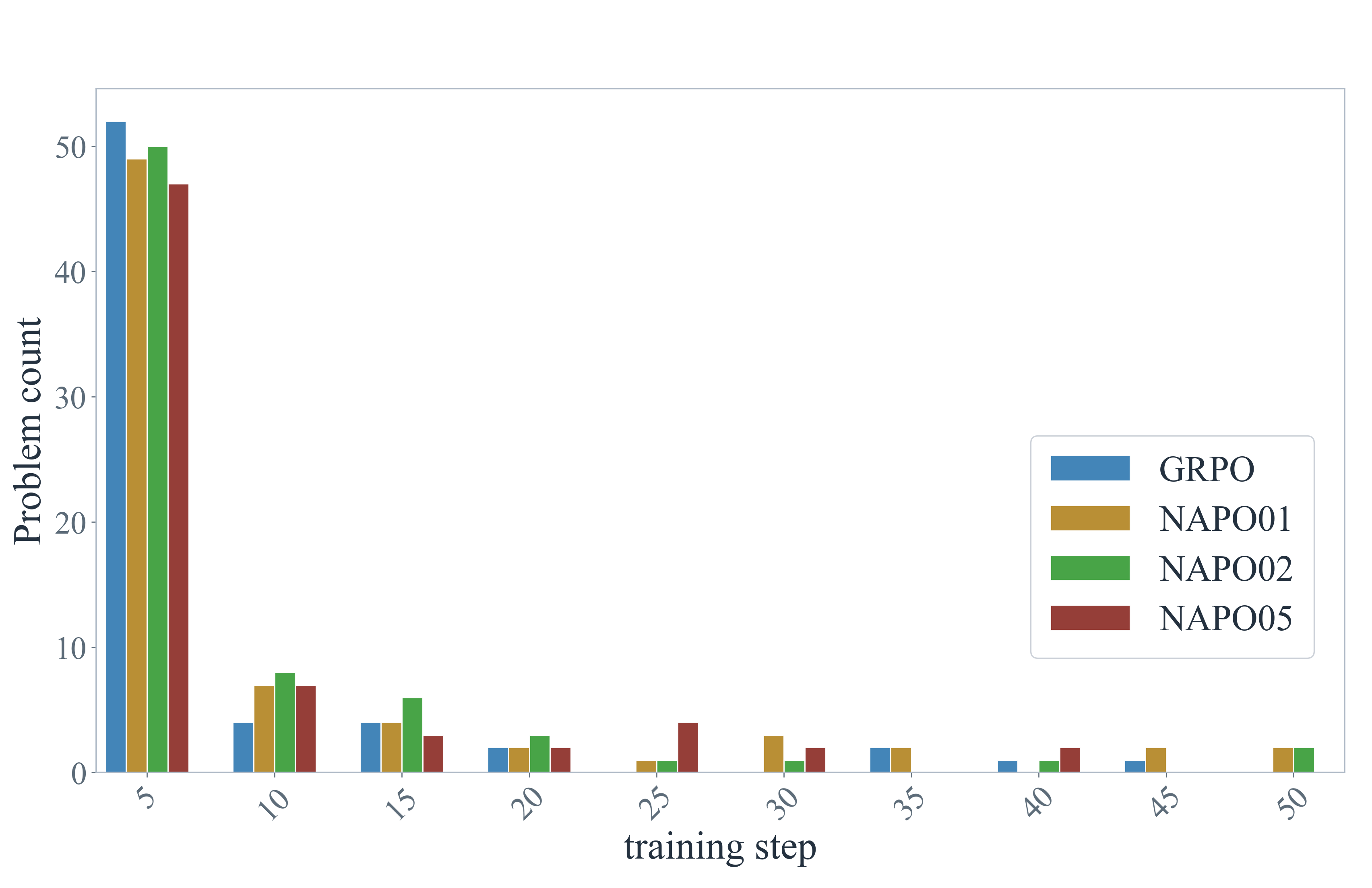}
        \caption{Number of solved problems across training steps. Setting the early fraction gate to 0.2 leads to the fastest exploration of correct solutions, consistent with our inference-time findings.}
        \label{fig:ng_training}
    \end{minipage}
\end{figure*}

\section{Algorithm}
\label{sec:analysis}
We now turn the empirical findings above into a training design. Our goal is to translate the driver into a practical update rule for RLVR.

\subsection{Theoretical Interpretation}
\label{sec:fisher_napo}

We provide a view of Fisher information to theoretically interpret the driver on the negative rollouts.

\paragraph{Fisher information as a measure of policy plasticity.}
Larger local Fisher means that a failed continuation is easier to reshape under a small parameter update, so the policy remains more \emph{plastic} around that rollout. Let $\mathbf{p}_t=\pi_\theta(\cdot\mid x,y_{<t})$ be the next-token distribution at step $t$, and let $\mathbf{F}_t$ denote the corresponding token-level Fisher contribution.

\begin{lemma}[Token-level Fisher information and entropy]
\label{lem:fisher_entropy_main}
\begingroup
\setlength{\abovedisplayskip}{2pt}
\setlength{\belowdisplayskip}{2pt}
\setlength{\abovedisplayshortskip}{1pt}
\setlength{\belowdisplayshortskip}{1pt}
For token distribution $\mathbf p_t$, the token-level Fisher trace is
\begin{equation*}
\mathrm{Tr}(\mathbf F_t)
=1-\sum_{k\in V}p_{t,k}^2.
\end{equation*}
Accordingly, Shannon entropy
serves as a useful proxy for local Fisher information: higher-entropy token distributions typically correspond to more policy-plastic states, although the two quantities are not identical in general.
\endgroup
\end{lemma}
\paragraph{Implication for negative learning.}
Lemma~\ref{lem:fisher_entropy_main} suggests that uniform penalties are wasteful: low-entropy failures are already rigid, so pushing them down mostly reinforces dead ends. By contrast, failures generated in higher-entropy states are more \emph{plastic} and retain nearby alternatives, making negative updates more likely to redirect the policy toward useful exploration. Empirically, the informative plasticity is concentrated early in the rollout, which is why NAPO uses the early-branching driver rather than generic response-level entropy. In short, we want to penalize failures more strongly where the policy is still movable.

\begin{tcolorbox}[colback=gray!5,colframe=black!70,title=\textbf{Takeaway},fonttitle=\bfseries]
The driver indicates that failures arising in more \emph{plastic} regions of the policy.
The N$\rightarrow$P phenomenon suggests that failures with smaller driver values are not random or hopeless: they come from prompts whose trajectories can cross into correctness under nearby policy shifts.

\end{tcolorbox}
This leads to a design intuition that larger penalties can be assigned to negative trajectories with larger drivers where the policy still retains meaningful uncertainty and plausible alternatives.

\begin{algorithm}[h]
\caption{NAPO: Negative-Augmented Policy Optimization}
\label{alg:napo}
\begin{algorithmic}[1]
\State Sample prompts $x$; for each $x$, sample rollouts $\{y_i\}$ and rewards $r(x,y_i)$;
\If{$\exists i: r(x,y_i)=+1$}
    \State Compute standard GRPO advantages $A_{\mathrm{GRPO}}(x,y_i)$;
    \State Update the policy with GRPO loss.
\Else
    \State Compute driver scores for all negative rollouts by \eqref{eq:early_ratio};
    \State Re-normalize scores into $[0,1]$ by \eqref{eq:driver_renorm};
    \State Assign rollout-level advantages by \eqref{eq:napo_adv_main};
    \State Update the policy with GRPO loss.
\EndIf
\end{algorithmic}
\end{algorithm}

\subsection{Negative-Augmented Policy Optimization}
\label{sec:napo_alg}

Motivated by the heterogeneity of negative rollouts, we propose \textbf{NAPO}, a GRPO-integrated algorithm that augments standard policy optimization with \emph{driver-weighted negative learning} on unsolved groups.
\paragraph{Driver-weighted negative advantage.}
NAPO uses the driver to modulate negative learning strength inside unsolved groups. Given driver scores $d(x,y_i)$ for rollouts in a fully negative group, we first re-normalize them into $[0,1]$ to avoid exploding gradient while preserving a relevant advantage magnitude:
\begin{equation}
\tilde d(x,y_i)
=
\frac{d(x,y_i)-\min_j d(x,y_j)}
{\max_j d(x,y_j)-\min_j d(x,y_j)+\epsilon}.
\label{eq:driver_renorm}
\end{equation}
We then assign rollout-level advantages as
\begin{equation}
A(x,y)=
\begin{cases}
A_{\mathrm{GRPO}}(x,y), & \text{if } p(x)=1,\\[4pt]
-\tilde d(x,y), & \text{if } p(x)=0,
\end{cases}
\label{eq:napo_adv_main}
\end{equation}
where $p(x)=\mathbb I[\exists i:r(x,y_i)=+1]$ indicates whether the current group is solved.

\paragraph{Objective: Grouped advantage on solved groups and weighted NSR on unsolved groups.}
Let
\(
p(x)=\mathbb I[\exists i:r(x,y_i)=+1]
\)
denote whether the current group is solved. Equivalently, the rollout-level advantage in Eq.~\eqref{eq:napo_adv_main} can be written as
\begin{equation}\label{eq:napo-adv-rollout}
\begin{aligned}
A(x,y_i)={}&p(x)\,A_{\mathrm{GRPO}}(x,y_i)\\
&-(1-p(x))\,\tilde d(x,y_i).
\end{aligned}
\end{equation}
When $p(x)=1$, Eq.~\eqref{eq:napo-adv-rollout} reduces to the standard GRPO clipped update.
When $p(x)=0$, we have $A(x,y_i)=-\tilde d(x,y_i)$, so the same clipped surrogate becomes a weighted NSR loss in which larger-driver negatives receive larger-magnitude penalties while clipping preserves the usual PPO/GRPO trust-region effect.

\section{Evaluation}
\label{sec:Z2O}

\begin{table*}[h]
\centering
\small
\setlength{\tabcolsep}{7pt}
\renewcommand{\arraystretch}{1.08}
\small
\begin{tabular}{lccccccccc}
\toprule
\textbf{Method} & \multicolumn{9}{c}{\textbf{Pass@k}} \\
\cmidrule(lr){2-10}
\textit{k} & 1 & 2 & 4 & 8 & 16 & 32 & 64 & 128 & 256 \\
\midrule

\multicolumn{10}{c}{\textbf{AMC23 }} \\
\midrule
Base & 0.2847 & 0.4512 & 0.6237 & 0.7546 & 0.8400 & \underline{0.9010} & \textbf{0.9495} & \textbf{0.9838} & \textbf{1.0000} \\
GRPO & \textbf{0.7352} & \textbf{0.7910} & \underline{0.8211} & \underline{0.8443} & \underline{0.8681} & 0.8972 & 0.9306 & 0.9579 & \underline{0.9750} \\
W-REINFORCE & 0.4514 & 0.5700 & 0.6570 & 0.7245 & 0.7900 & 0.8555 & 0.9086 & 0.9450 & \underline{0.9750} \\
\rowcolor{gray!20}
NAPO02 & 0.7141 & 0.7718 & 0.8051 & 0.8328 & 0.8628 & 0.8940 & 0.9222 & 0.9467 & \underline{0.9750} \\
\rowcolor{gray!20}
NAPO05 & \underline{0.7191} & \underline{0.7828} & \textbf{0.8247} & \textbf{0.8556} & \textbf{0.8845} & \textbf{0.9147} & \underline{0.9438} & \underline{0.9654} & \underline{0.9750} \\
\midrule

\multicolumn{10}{c}{\textbf{AIME2025 }} \\
\midrule
Base & 0.0305 & 0.0561 & 0.0964 & 0.1496 & 0.2074 & 0.2657 & 0.3216 & 0.3684 & 0.4000 \\
GRPO & \textbf{0.1346} & \textbf{0.1753} & \textbf{0.2165} & \underline{0.2563} & 0.2910 & 0.3271 & 0.3740 & 0.4294 & \underline{0.4667} \\
W-REINFORCE & 0.0254 & 0.0439 & 0.0689 & 0.0970 & 0.1300 & 0.1767 & 0.2425 & 0.3278 & 0.4333 \\
\rowcolor{gray!20}
NAPO02 & \underline{0.1220} & \underline{0.1648} & \underline{0.2113} & \textbf{0.2631} & \textbf{0.3175} & \textbf{0.3695} & \textbf{0.4195} & \textbf{0.4745} & \textbf{0.5333} \\
\rowcolor{gray!20}
NAPO05 & 0.1138 & 0.1576 & 0.2008 & 0.2475 & \underline{0.3006} & \underline{0.3555} & \underline{0.4130} & \underline{0.4741} & \textbf{0.5333} \\
\midrule

\multicolumn{10}{c}{\textbf{AIME24}} \\
\midrule
Base & 0.0857 & 0.1508 & 0.2415 & 0.3425 & 0.4343 & 0.5112 & 0.5819 & 0.6575 & 0.7333 \\
GRPO & 0.2832 & 0.3574 & 0.4242 & 0.4860 & 0.5391 & 0.5869 & 0.6330 & 0.6719 & 0.7000 \\
W-REINFORCE & 0.1516 & 0.2322 & 0.3105 & 0.3706 & 0.4152 & 0.4500 & 0.4821 & 0.5209 & 0.5667 \\
\rowcolor{gray!20}
NAPO02 & \textbf{0.2988} & \textbf{0.3878} & \textbf{0.4648} & \textbf{0.5314} & \textbf{0.5917} & \textbf{0.6504} & \textbf{0.7048} & \textbf{0.7564} & \textbf{0.8000} \\
\rowcolor{gray!20}
NAPO05 & \underline{0.2898} & \underline{0.3762} & \underline{0.4490} & \underline{0.5133} & \underline{0.5734} & \underline{0.6301} & \underline{0.6810} & \underline{0.7248} & \underline{0.7667} \\
\midrule

\multicolumn{10}{c}{\textbf{Math-Beyond}} \\
\midrule
Base & 0.0426 & 0.0702 & 0.1051 & 0.1446 & 0.1879 & 0.2357 & \textbf{0.2866} & \textbf{0.3391}& \textbf{0.3978} \\
GRPO & 0.1337 & 0.1574 & 0.1780 & 0.1984 & 0.2205 & 0.2454 & 0.2723 & 0.3014 & 0.3370 \\
W-REINFORCE & 0.0587 & 0.0866 & 0.1147 & 0.1400 & 0.1639 & 0.1899 & 0.2197 & 0.2564 & 0.3094 \\
\rowcolor{gray!20}
NAPO02 & \underline{0.1348} & \underline{0.1612} & \underline{0.1882} & \underline{0.2130} & \underline{0.2334} & \textbf{0.2527} & \underline{0.2763} & \underline{0.3105} & \underline{0.3591} \\
\rowcolor{gray!20}
NAPO05 & \textbf{0.1383} & \textbf{0.1680} & \textbf{0.1948} & \textbf{0.2164} & \textbf{0.2337} & \underline{0.2508} & 0.2704 & 0.2926 & 0.3204 \\
\bottomrule
\end{tabular}

\caption{Pass@k comparison across datasets. The row corresponding to our method is highlighted in gray, the best value within each dataset block and column is shown in bold, and the second-best value is underlined. For W-REINFORCE method, we set the positive advantage weight to 0.5 to handle the much harder prompts.}
\label{tab:passk_main}
\end{table*}

We empirically evaluate the effectiveness of NAPO on hard prompts.

\paragraph{Setup.}
We evaluate NAPO under a fixed rollout budget and compare against GRPO and negative-sample-based baselines, including NSR's weighted variant (w-REINFORCE).
We also evaluate the impact of early fraction ratio, where NAPO02 refers to taking the first $20\%$ of the response in the driver, while NAPO05 takes $50\%$ and NAPO01 takes $10\%$.
We report (i) Pass@k on validation sets and (ii) TTFP.
Experiments are conducted on multiple validation sets (AMC23, AIME2025, AIME24, and Math-Beyond), covering a wide range of difficulty levels.
\footnote{We additionally report training-batch transition statistics, with extended ablations deferred to the appendix.}

\paragraph{Training protocol.}
We use the verl framework \citep{sheng2025hybridflow} and train on DAPO-17k \cite{yu2025dapoopensourcellmreinforcement}, a challenging dataset for Qwen2.5-Math-7B \cite{yang2024qwen2}, where the initial pass rate is below $0.2$.
For each prompt, we sample $G=8$ rollouts per step.
Unless otherwise specified, we use temperature $T=1.0$ during training, a maximum context length of 4096, and train for one epoch.\footnote{All other hyperparameters follow Section~\ref{sec:setup}, with full details provided in Appendix~\ref{D.1.} and Table~\ref{tab:hyperparams_grpo}.}
We report both large-scale validation results and controlled experiments under identical training batches (as in \S~\ref{sec:findings}).

\paragraph{Results.}

\textbf{(1) Improved sample efficiency on hard prompts.}
As shown in Figure~\ref{fig:ttfp_nvsg}, NAPO discovers first-positive rollouts earlier across datasets, solving more prompts during early training under the same rollout budget.
This indicates improved sample efficiency, particularly in low-success regimes where standard GRPO suffers from weak learning signals.
These gains require little additional computation: NAPO uses no extra rollout, verifier call, auxiliary model, or policy forward pass.
We compute the uncertainty statistic from logits already produced during the old-policy forward pass and retain one scalar per valid response token rather than a vocabulary-sized tensor; the remaining operations are masking, sequence-wise reductions, and within-group min--max normalization.
Across common logged steps, the advantage calculation time is $0.935 \pm 0.030\,\mathrm{s/step}$ for NAPO and $0.525 \pm 0.016\,\mathrm{s/step}$ for GRPO (mean $\pm$ standard deviation).
The resulting increment is only \textbf{0.410 s/step}, approximately \textbf{0.03\%} of the observed end-to-end step time.
Thus, the trace shows no material training-time overhead.

\textbf{(2) Effect of negative scaling.}
Comparing NAPO and NAPO05 reveals a clear trade-off rather than a uniform winner.
The milder NAPO05 variant performs best on AMC23, on the highest-$k$ region of AIME2025, and on low-$k$ Math-Beyond, whereas full NAPO is clearly strongest on AIME24 and in the mid-$k$ regime of AIME2025.
This pattern suggests that negative updates should be tuned to benchmark difficulty: stronger shaping helps when there is substantial room for zero-to-one improvement, while milder scaling is more robust once exploration is less constrained.

\textbf{(3) Within-group differentiation versus uniform negative feedback.}
To isolate the effect of within-group differentiation, we add a Uniform($-0.1$) control that assigns the same small negative advantage to every rollout in an all-negative group.
All training and evaluation settings are otherwise matched.
Table~\ref{tab:uniform-control} reports the mean paired improvement of NAPO over this control across four independent evaluations.

\begin{table}[t]
\centering
\small
\setlength{\tabcolsep}{4pt}
\begin{tabular}{@{}lrrr@{}}
\toprule
& \multicolumn{3}{c}{$\Delta\mathrm{Pass}@k$ (\%)} \\
\cmidrule(lr){2-4}
Dataset & $k=1$ & $k=8$ & $k=256$ \\
\midrule
AMC23       & $+14.40$ & $+7.95$  & $+1.88$ \\
AIME2025    & $+10.48$ & $+15.31$ & $+8.33$ \\
AIME24      & $+20.15$ & $+27.28$ & $+17.50$ \\
Math-Beyond & $+7.43$  & $+8.73$  & $+0.41$ \\
\bottomrule
\end{tabular}
\caption{$\mathrm{Pass}@k$ gains of NAPO over Uniform($-0.1$) across datasets at $T=1.0$, averaged over four rollout reevaluations with 256 generations per question. Values are percentage points; positive values favor NAPO.}
\label{tab:uniform-control}
\end{table}

Among the evaluation results above, the separation is strongest on the AIME benchmarks---for example, $+27.28$ points at $\mathrm{Pass}@8$ on AIME24---and the mean remains positive at $\mathrm{Pass}@256$ on all four datasets.
In a separate single-evaluation control with Uniform($-1$), NAPO improves $\mathrm{Pass}@256$ by $17.50$, $23.33$, $30.00$, and $17.13$ percentage points on AMC23, AIME2025, AIME24, and Math-Beyond, respectively.
Thus, simply activating constant negative feedback does not reproduce NAPO in these runs; these controls isolate the benefit of driver-based within-group differentiation over a constant update.
Because the four reevaluations use fixed checkpoints rather than independent training runs, they establish evaluation-level consistency but not training-seed significance.

\section{Conclusion}
We studied the zero-to-one problem in RLVR, motivated by two obstacles: diagnosing $\mathrm{N}\!\to\!\mathrm{P}$ transitions under binary supervision, and extracting informative gradients from heterogeneous negatives in fully-unsolved groups. 
Using event-based metrics (TTFP and $\mathrm{N}\!\to\!\mathrm{P}$ events), we showed that failures preceding successful transitions retain higher early-stage uncertainty, providing a practical signal to separate near-solvable rollouts from dead ends. 
Building on this, we proposed \textsc{NAPO}, a GRPO-integrated method that assigns graded negative advantages according to driver scores under a Fisher-information view, thereby injecting structured gradients into unsolved groups while preserving stable updates on solved ones. 
\textsc{NAPO} accelerates early solution discovery and improves Pass@256 by $10.0$ points on AIME24 and $6.6$ points on AIME2025 compared with GRPO, confirming that driver-guided negative learning is an effective way to broaden reasoning boundaries in low-success RLVR.

\section*{Limitations}

Our analysis and method are currently focused on math-oriented RLVR settings with binary verifiers. Although these tasks provide a clean testbed for studying zero-to-one transitions, the same entropy-based driver may behave differently in broader domains, longer-horizon reasoning tasks, or environments with partial-credit, noisy, or subjective feedback. Extending NAPO to such settings may require more robust driver definitions and calibration procedures.

NAPO also relies on rollout-level statistics collected during training. Its benefits may therefore depend on the rollout budget, group size, decoding temperature, and the quality of the base policy. When the model rarely approaches a correct solution, early uncertainty patterns may be insufficient to identify useful negative samples; conversely, when success rates are already high, the additional weighting of negatives may provide limited gains. Future work should evaluate the method under more diverse model scales, verifier designs, and compute budgets.

\section*{Acknowledgments}
\thanks{This work was supported in part by China NSF grant No.U2268204, 62322206, 62432007, 62441236, 62132018, 62025204, 62272307, 62372296, U25A6024. The opinions, findings, conclusions, and recommendations expressed in this paper are those of the authors and do not necessarily reflect the views of the funding agencies or the government.}

\bibliography{custom,anthology-1,anthology-2}
\newpage
\appendix
\section{Experiment Settings}
\label{D.1.}
\subsection{Chat template}

\begin{tcolorbox}[colback=gray!5,colframe=black!70,title=\textbf{Chat template for Qwen},fonttitle=\bfseries]
\small
\textbf{System:} \texttt{You are a helpful assistant.}

\vspace{0.35em}
\textbf{User:} \texttt{\{input\} Please reason step by step, and put your final answer within \textbackslash boxed\{\}.}

\vspace{0.35em}
\textbf{Assistant:} \texttt{Model generates a step-by-step solution; the final answer is enclosed in \textbackslash boxed\{\}.}
\end{tcolorbox}

\subsection{Hyperparameter Settings}
\begin{table}[ht]
\centering
\small
\setlength{\tabcolsep}{7pt}
\resizebox{\linewidth}{!}{%
\begin{tabular}{ll}
\toprule
\textbf{Category} & \textbf{Setting} \\
\midrule
\multicolumn{2}{l}{\textit{Model}}\\
Model & \texttt{Qwen2.5-Math-7B} \\
\midrule
\multicolumn{2}{l}{\textit{Data}}\\
Train batch size & 1024 \\
Max prompt length & 1024  \\
Max response length & 3072  \\
\midrule

\multicolumn{2}{l}{\textit{RL Algorithm }}\\
KL coefficient & 0.0 \\
Entropy coefficient & 0.0  \\
Critic warmup & 0  \\
\midrule

\multicolumn{2}{l}{\textit{Optimization}}\\
Learning rate & $1\times10^{-6}$ \\
PPO minibatch size & 256 \\
\midrule
\multicolumn{2}{l}
{\textit{Sampling}} \\
top\_k & -1 \\
top\_p & 1 \\
temperature & 1.0 \\
\bottomrule
\end{tabular}%
}

\caption{Detailed training hyperparameters.}
\label{tab:hyperparams_grpo}
\end{table}
All experiments are conducted on a single node equipped with eight NVIDIA H100 GPUs. 
We report final evaluation results using the checkpoint at step 200 for Qwen2.5-Math-7B and the checkpoint at step 100 for Llama-3.1-8B-Instruct.
\section{Definitions of Batch-Level Driver Statistics}
\label{app:driver_stats}

In this appendix, we define the batch-level summary statistics used in Table~\ref{tab:driver_batch_stats_appendix} and Figure~\ref{fig:early_ppl_inbatch} to characterize the distribution of rollout-time drivers within the N$\rightarrow$P group. For Figure~\ref{fig:early_ppl_inbatch}
We use kernel density estimation (KDE) only for visualization, to provide a smoothed view of the empirical distribution of the driver scores. 
KDE is a standard non-parametric density estimation method that places a smooth kernel around each observed sample and aggregates these kernels to estimate a continuous density. 
It does not affect training, evaluation, or the computation of the proposed driver; all quantitative analyses are performed on the original empirical scores.

Let $\{d_i\}_{i=1}^{n}$ denote the driver values computed for the rollouts in a given subset (e.g., the N$\rightarrow$P rollout group of a batch), where $n$ is the number of rollouts in that subset.

\paragraph{Mean-shift statistic (\texttt{msk}).}
The statistic \texttt{msk} summarizes the average magnitude of the driver values within the selected rollout group:
\begin{equation}
\texttt{msk}
=
\frac{1}{n}\sum_{i=1}^{n} d_i.
\end{equation}
In our analysis, larger \texttt{msk} indicates that the corresponding batch tends to allocate relatively more uncertainty to the early stage of generation.

\paragraph{Right-tail mass statistic (\texttt{rt}).}
The statistic \texttt{rt} summarizes the heaviness of the right tail of the driver distribution. Let $\tau$ be a fixed upper-tail threshold (we set $\tau=1.95$ in practice). We define
\begin{equation}
\texttt{rt}
=
\frac{1}{n}\sum_{i=1}^{n}\mathbf{1}[d_i > \tau],
\end{equation}
where $\mathbf{1}[\cdot]$ is the indicator function. Larger \texttt{rt} indicates that a larger fraction of rollouts fall into the high-driver region, i.e., the distribution has a heavier right tail.

\paragraph{Interpretation.}
These statistics are not intended to serve as calibrated predictors of future success or as exact estimates of future reward. In the main text, we use them only to support a weaker but more practically relevant claim: larger driver values, especially a heavier right tail, are enriched in rollout groups associated with stronger N$\rightarrow$P tendency. This is the only property required by our method design, where the driver is used as a surrogate ordering signal rather than as a direct estimator of future pass rate.

\begin{table}[tbp]
\centering
\small
\setlength{\tabcolsep}{8pt}
\resizebox{\linewidth}{!}{%
\begin{tabular}{lcccc}
\toprule
Replay & Training N$\rightarrow$P & \texttt{msk} & \texttt{rt} \\
\midrule
\textit{n1} & 120 & 0.924 & 0.039 \\
\textit{n2} & 117.67 & 0.812 & 0.026 \\
\bottomrule
\end{tabular}%
}
\caption{Representative batch-level driver statistics for a good(n1)/bad(n2) replay comparison. Larger \texttt{msk} and \texttt{rt} indicate more mass in the high-driver region.}
\label{tab:driver_batch_stats_appendix}
\end{table}

\section{Additional results}
\label{app:addition_res}
\begin{figure*}[tbp]
    \centering
    \includegraphics[width=1.0\linewidth]{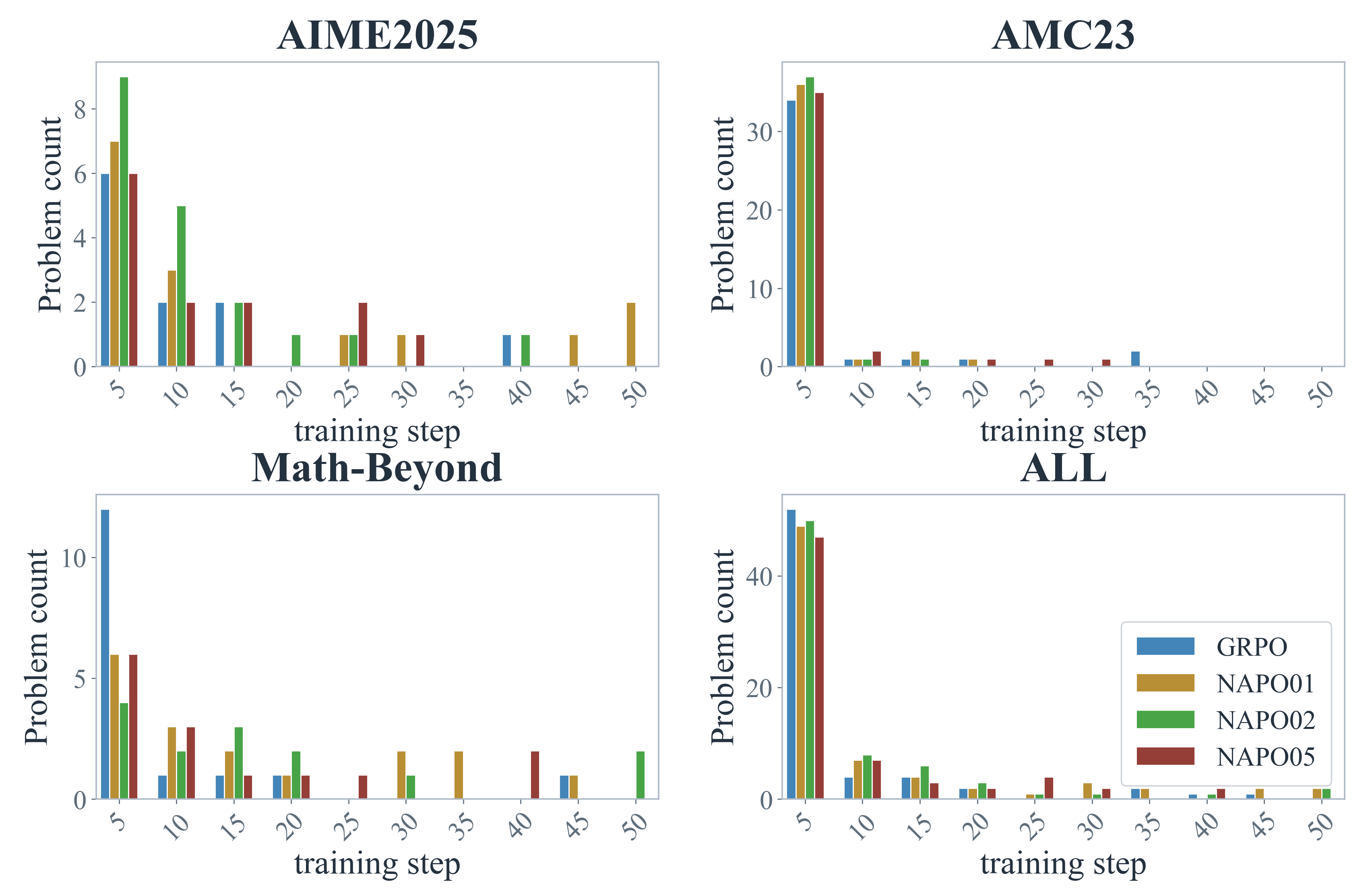}
    \caption{TTFP distribution}
    \label{fig:cumu-solved-num}
\end{figure*}
\begin{figure*}[tbp]
    \centering
    \includegraphics[width=1.0\linewidth]{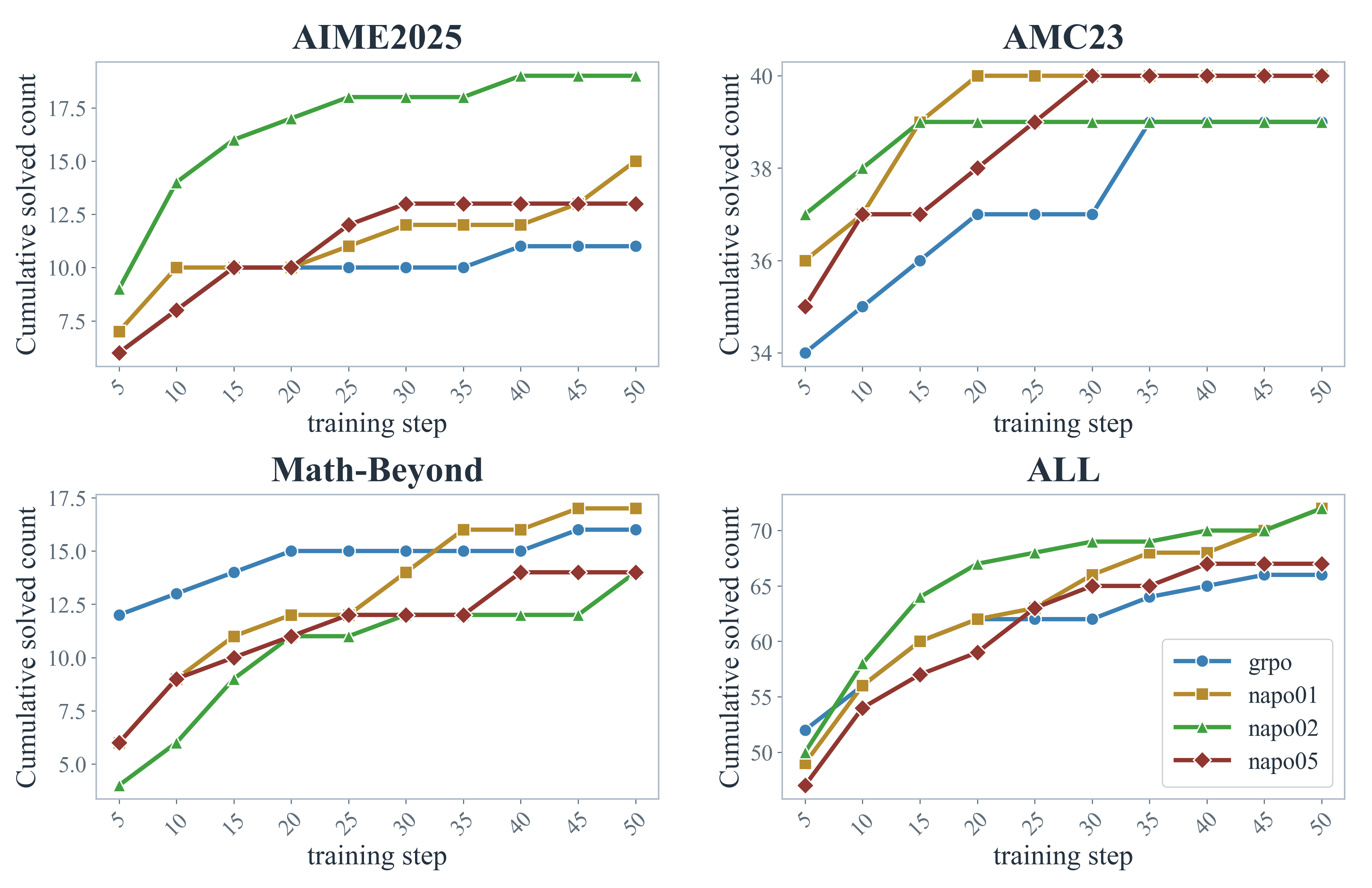}
    \caption{Cumulative solved number}
    \label{fig:cumu-solved-num-detail}
\end{figure*}
\subsection{Generalization beyond Qwen2.5-Math-7B}
\label{app:llama_generalization}

\paragraph{Transition-level evidence.}
Starting from a shared Llama-3.1-8B-Instruct checkpoint, we run five stochastic GRPO continuations.
Across the 696 prompts that are negative at the shared checkpoint, the mean N$\to$P rate is $(14.86 \pm 0.44)\%$ (mean $\pm$ standard error across continuations).
Thus, delayed first-solution discovery is not confined to Qwen2.5-Math-7B.

\paragraph{Training-level evidence.}
We apply NAPO using the same early-fraction window $\alpha=0.2$ selected on Qwen2.5-Math-7B (Figure~\ref{fig:ng_training}), without Llama-specific tuning.
Table~\ref{tab:passk_llama} compares the resulting step-100 checkpoint against GRPO.

\begin{table*}[t]
\centering
\small
\setlength{\tabcolsep}{7pt}
\renewcommand{\arraystretch}{1.08}
\begin{tabular}{lcccc}
\toprule
\textbf{Method (Llama)} & \multicolumn{4}{c}{\textbf{Pass@k (\%)}} \\
\cmidrule(lr){2-5}
\textit{k} & 1 & 2 & 8 & 256 \\
\midrule

\multicolumn{5}{c}{\textbf{AMC23}} \\
\midrule
\rowcolor{gray!20}
NAPO02 & \textbf{40.50} & \textbf{47.68} & 57.82 & \textbf{80.83} \\
GRPO   & 37.55 & 47.29 & \textbf{59.89} & 80.00 \\
\midrule

\multicolumn{5}{c}{\textbf{AIME2025}} \\
\midrule
\rowcolor{gray!20}
NAPO02 & 0.06 & 0.11 & 0.45 & 13.33 \\
GRPO   & \textbf{0.17} & \textbf{0.33} & \textbf{1.25} & \textbf{20.00} \\
\midrule

\multicolumn{5}{c}{\textbf{AIME24}} \\
\midrule
\rowcolor{gray!20}
NAPO02 & 4.75 & 6.67 & 11.96 & \textbf{37.78} \\
GRPO   & \textbf{7.11} & \textbf{10.41} & \textbf{16.15} & 33.33 \\
\midrule

\multicolumn{5}{c}{\textbf{Math-Beyond}} \\
\midrule
\rowcolor{gray!20}
NAPO02 & \textbf{5.90} & \textbf{7.32} & \textbf{9.68} & \textbf{17.50} \\
GRPO   & 4.93 & 6.39 & 8.92 & 16.21 \\
\bottomrule
\end{tabular}

\caption{Pass@k comparison across datasets for \textbf{Llama-3.1-8B-Instruct} at temperature $T=1.0$. Values are percentages. The row corresponding to NAPO02 is highlighted in gray, and the best value between GRPO and NAPO02 within each dataset block and column is shown in bold. Due to limited computational resources, we evaluate the checkpoint at step 100.}
\label{tab:passk_llama}
\end{table*}

The results are mixed, where NAPO performs best on Math-Beyond and high-budget AIME24, while degrading on AIME2025. Thus, the evaluation indicates that the driver window and resulting advantage magnitudes may require calibration across model families, while remaining robust to a certain degree.

\subsection{Beyond mathematics: binary-verifiable code}
\label{app:code_generalization}

\paragraph{Setup.}
We examine whether the transition-level entropy signal extends beyond mathematical reasoning.
We train Qwen3-4B-Instruct~\citep{yang2025qwen3technicalreport} on 512 distinct coding problems from LiveCodeBench v5~\citep{LCB} using execution-based binary rewards, and evaluate on 100 held-out problems from LiveCodeBench v6.
Five consecutive GRPO steps each visit all 512 training prompts and sample $G=8$ rollouts per prompt, yielding four adjacent transitions for the analysis below.
All other settings follow Section~\ref{sec:setup}.

\paragraph{Code-domain entropy statistic.}
On the fixed sensitivity grid $\alpha\in\{0.1,0.2,0.5\}$ evaluated on the same transitions, $\alpha=0.5$ gives the clearest observed separation, consistent with the different output structure and prompt formatting of coding tasks.
For this experiment, \texttt{msk} denotes the prompt-level mean early-to-global plain-entropy ratio across the eight source-step rollouts.
Unlike the NAPO training driver $d_{\text{early-ppl}}$, this analysis-only statistic uses entropy directly rather than its exponential.

\begin{table*}[t]
\centering
\small
\setlength{\tabcolsep}{7pt}
\renewcommand{\arraystretch}{1.08}
\begin{tabular}{lrrrr}
\toprule
Transition & N$\to$P / N$\to$N & \texttt{msk} N$\to$P & \texttt{msk} N$\to$N & $\Delta\texttt{msk}$ \\
\midrule
$1\to2$ & $20/133$ & 1.1044 & 1.0740 & $+0.0304$ \\
$2\to3$ & $20/135$ & 1.0997 & 1.0710 & $+0.0287$ \\
$3\to4$ & $18/144$ & 1.0968 & 1.0679 & $+0.0289$ \\
$4\to5$ & $25/138$ & 1.0941 & 1.0710 & $+0.0231$ \\
\bottomrule
\end{tabular}
\caption{Transition-level entropy statistics for \textbf{Qwen3-4B-Instruct} on LiveCodeBench at $\alpha=0.5$. The second column reports the N$\to$P and N$\to$N prompt counts, and the remaining columns report their mean \texttt{msk} values and difference; positive differences indicate higher early relative entropy for N$\to$P.}
\label{tab:code_transition_entropy}
\end{table*}

Pooling the four transitions gives 83 N$\to$P and 550 N$\to$N prompt--transition observations, whose mean \texttt{msk} values are 1.0985 and 1.0709, respectively.
The difference is $+0.0276$, with a 95\% prompt-clustered bootstrap confidence interval of $[0.0066,0.0484]$.
The step-stratified AUC is $0.608$, with a 95\% confidence interval of $[0.532,0.681]$.
The difference has the same sign at all four transitions rather than being driven by a single favorable step.
These results provide preliminary evidence for an entropy--exploration signal in the coding tasks.

\section{Theoretical Derivation: Fisher Information and Negative-Augmented Policy Optimization}
\label{app:Fisher}
In this section, we provide a theoretical justification for the proposed driver-based
weighting scheme.  We demonstrate that penalizing negative samples with uniform
advantage (standard GRPO) wastes optimization budget on samples that provide
negligible Fisher Information.  We then derive Negative-Augmented Policy
Optimization (NAPO), where the advantage is proportional to the Fisher
Information of the trajectory, effectively maximizing the information gain of the
policy update.

\subsection{Fisher Information of Negative Samples}

Consider a prompt $x$ with a rollout group
$\mathcal{G}^{-}=\{y_{i}\}_{i=1}^{G}$ where all rollouts are negative, i.e.,
$r(x,y_{i})=-1$.  In policy optimization, the Fisher Information Matrix (FIM)
$\mathbf{F}(\theta)$ measures the curvature of the KL divergence and the
informativeness of a sample gradient.  For a specific trajectory $y$, the
contribution to the FIM is given by the outer product of the score function:
\begin{equation}\label{eq:fim-sample}
  \mathbf{F}_{y}(\theta)
  = \nabla_{\theta}\log\pi_{\theta}(y|x)\,
    \nabla_{\theta}\log\pi_{\theta}(y|x)^{\top}.
  \tag{32}
\end{equation}
A sample $y$ is considered informative for optimization if its gradient vector
has a large magnitude in the parameter space, indicating a significant potential
to reshape the policy distribution.

We analyze the Fisher Information at the token level.  Let
$\mathbf{p}_{t}=\pi_{\theta}(\cdot|x,y_{<t})$ be the probability vector over
the vocabulary $V$ at step $t$.  Let $\mathbf{F}_{t}$ denote the Fisher
Information contribution from position $t$.

\begin{lemma}[Fisher Information and Entropy]\label{lem:fisher-entropy}
Let $H_{t}=-\sum_{k}p_{k}\log p_{k}$ be the entropy of the distribution at
step $t$.  The trace of the token-level Fisher Information matrix satisfies:
\begin{equation}\label{eq:trace-fisher}
  \operatorname{Tr}(\mathbf{F}_{t})
  = 1 - \sum_{k\in V} p_{k}^{2}.
  \tag{33}
\end{equation}
Furthermore, for a fixed vocabulary size $V$, the quantity
$1-\sum p_{k}^{2}$ and the entropy $H_{t}$ are positively associated:
both are maximized by the uniform distribution (at $\tfrac{V-1}{V}$ and
$\log V$, respectively) and both equal zero only for deterministic
distributions.  In particular, $H_{t}\to 0$ implies
$\operatorname{Tr}(\mathbf{F}_{t})\to 0$.
\end{lemma}

\begin{proof}
The score function for a sampled token $y_{t}$ is
$\nabla_{\theta}\log\pi_{\theta}(y_{t}|x,y_{<t})$.
Under the common softmax parameterization
$p_{k}=\frac{e^{z_{k}}}{\sum e^{z_{k}}}$,
the gradient of the log-probability with respect to the logits $\mathbf{z}$ is:
\begin{equation}\label{eq:score}
  \nabla_{\mathbf{z}}\log p(y_{t})
  = \mathbf{e}_{y_{t}} - \mathbf{p}_{t},
  \tag{34}
\end{equation}
where $\mathbf{e}_{y_{t}}$ is the one-hot vector for the sampled token.
The Fisher Information matrix is the covariance of this gradient:
\begin{align}
  \mathbf{F}_{t}
  &= \mathbb{E}_{y_{t}}\bigl[
       (\mathbf{e}_{y_{t}}-\mathbf{p}_{t})
       (\mathbf{e}_{y_{t}}-\mathbf{p}_{t})^{\top}
     \bigr]  \tag{35} \\
  &= \mathbb{E}[\mathbf{e}_{y_{t}}\mathbf{e}_{y_{t}}^{\top}]
     - \mathbf{p}_{t}\mathbf{p}_{t}^{\top}.  \tag{36}
\end{align}
Since
$\mathbb{E}[\mathbf{e}_{y_{t}}\mathbf{e}_{y_{t}}^{\top}]
 =\operatorname{diag}(\mathbf{p}_{t})$,
the trace of the FIM is:
\begin{align}
  \operatorname{Tr}(\mathbf{F}_{t})
  &= \operatorname{Tr}\bigl(\operatorname{diag}(\mathbf{p}_{t})\bigr)
     - \operatorname{Tr}(\mathbf{p}_{t}\mathbf{p}_{t}^{\top})  \tag{37} \\
  &= \sum_{k}p_{k} - \sum_{k}p_{k}^{2}
   = 1 - \sum_{k}p_{k}^{2}.  \tag{38}
\end{align}

The quantity $G := 1-\sum p_{k}^{2}$ is the Gini impurity (equivalently, the
Tsallis entropy of order~2).  Both $G$ and the Shannon entropy
$H=-\sum p_{k}\log p_{k}$ are \emph{Schur-concave} functions of the
probability vector: they are both uniquely maximized by the uniform distribution
and both equal zero if and only if the distribution is deterministic.
While $G$ and $H$ are not in general strict monotonic functions of each other
for $V\ge 3$, they are tightly coupled via the R\'{e}nyi entropy ordering
$H_{1}\ge H_{2}$, which gives:
\[
  1-\sum_{k}p_{k}^{2}
  \;\le\; 1-e^{-H}.
\]
In particular, $H\to 0$ forces $\operatorname{Tr}(\mathbf{F}_{t})\to 0$, and
conversely the uniform distribution uniquely maximizes both quantities.
Hence, higher-entropy distributions correspond, in a strong qualitative sense,
to larger local Fisher information.
\end{proof}

\subsection{Analysis of Standard NSR on Unsolved Groups}

Standard NSR assigns uniform negative advantages $A(y)\approx -1$ to all
samples in an unsolved group.  From the perspective of Fisher Information, this
implies that the optimization update treats samples with zero information
(confident errors) and high information (uncertain errors) equally.

\begin{proposition}[Gradient Saturation and Fisher Information]\label{prop:saturation}
Consider a negative sample $y$ generated from a low-entropy (confident) policy
state.
\begin{enumerate}
\item \textbf{Zero Information:}
  If $H_{t}\to 0$, then $\operatorname{Tr}(\mathbf{F}_{t})\to 0$.
  The gradient $\nabla_{\theta}\log\pi_{\theta}(y_{t})$ vanishes.
\item \textbf{Saddle Points:}
  Negative samples generated with high confidence correspond to ``saddle
  points'' or ``dead ends'' in the likelihood landscape.  Uniform
  penalization ($A=-1$) attempts to push probability mass away from
  these points, but the vanishing Fisher Information implies that
  $\Delta\theta\approx 0$ regardless of the advantage magnitude.
\end{enumerate}
Thus, uniform advantage assignment wastes gradient budget on samples that lie in
low-information regions of the parameter space.
\end{proposition}

\begin{proof}
The gradient vector for the sampled token $y_{t}$ is
$\mathbf{g}_{t}=\mathbf{e}_{y_{t}}-\mathbf{p}_{t}$, where
$\mathbf{e}_{y_{t}}$ is the one-hot encoding of the sampled token and
$\mathbf{p}_{t}$ is the probability vector $\pi_{\theta}(\cdot|x,y_{<t})$.
The squared norm of this gradient is:
\begin{align}
  \|\mathbf{g}_{t}\|_{2}^{2}
  &= \|\mathbf{e}_{y_{t}}-\mathbf{p}_{t}\|_{2}^{2}  \tag{39} \\
  &= (1-\pi(y_{t}))^{2}+\sum_{v\neq y_{t}}\pi(v)^{2}.  \tag{40}
\end{align}
As $\pi(y_{t})\to 1$ (confidence approaches certainty, entropy
$H_{t}\to 0$), the first term approaches~0.  Simultaneously, the mass of the
probability vector concentrates on $y_{t}$, forcing $\pi(v)\to 0$ for all
other $v$.  Thus, $\|\mathbf{g}_{t}\|_{2}\to 0$.

This implies that confidently generated tokens (which characterize the
``dead-end'' negative rollouts in Section~\ref{sec:early-branch}) contribute negligible gradient
signal.  Uniformly weighting such samples wastes optimization budget on
saturated gradients that cannot effectively redirect the policy.
\end{proof}

\subsection{Derivation of NAPO}\label{sec:napo-deriv}

To maximize the efficiency of the update, we should allocate the ``negative
learning budget'' to the samples whose penalization best reduces the total
likelihood of failed trajectories.  We formalize this as a constrained
optimization over per-rollout weights and show that the solution naturally
connects to Fisher Information and, through it, to trajectory entropy.

\paragraph{Exact formulation via the Gram matrix.}
Let $\mathcal{G}^{-}=\{y_i\}_{i=1}^{G}$ be a fully negative rollout group for
prompt~$x$.  Write
$g_{y}=\nabla_{\theta}\log\pi_{\theta}(y\mid x)$ for the trajectory-level
score vector, and define the \emph{Gram matrix}
$\mathbf{G}\in\mathbb{R}^{G\times G}$ with entries
\begin{equation}\label{eq:gram}
  G_{ij} = g_{y_i}^{\!\top}\, g_{y_j}.
  \tag{43}
\end{equation}

Given a weight vector $w=(w_1,\dots,w_G)^{\!\top}$, the parameter update is
$\delta=-\eta\sum_{i}w_i\,g_{y_i}$.  The total first-order reduction in the
log-likelihood of every negative rollout in the group is
\begin{equation}\label{eq:total-reduction}
  R(w)
  \;=\;-\sum_{j=1}^{G} g_{y_j}^{\!\top}\delta
  \;=\;\eta\;\mathbf{1}^{\!\top}\mathbf{G}\,w.
  \tag{44}
\end{equation}

We seek the weight vector that maximizes $R(w)$ subject to a variance budget
$\|w\|^{2}\le B$ (which controls the variance of the stochastic gradient
and ensures stable updates):
\begin{equation}\label{eq:opt-problem}
  \max_{w}\;\mathbf{1}^{\!\top}\mathbf{G}\,w
  \qquad\text{s.t.}\qquad \|w\|^{2}\le B.
  \tag{45}
\end{equation}

By the Cauchy--Schwarz inequality,
$\mathbf{1}^{\!\top}\mathbf{G}\,w
 \le \|\mathbf{G}^{\!\top}\mathbf{1}\|\;\|w\|$,
with equality when $w\propto\mathbf{G}^{\!\top}\mathbf{1}$.
Since $\mathbf{G}$ is symmetric, the exact solution is
\begin{equation}\label{eq:exact-w}
\begin{aligned}
w^{*} &\propto \mathbf{G}\,\mathbf{1},\\
w^{*}(y_i)
&\propto \bigl[\mathbf{G}\,\mathbf{1}\bigr]_{i}
=\sum_{j=1}^{G} g_{y_i}^{\!\top}\,g_{y_j}.
\end{aligned}
\tag{46}
\end{equation}

Computing $\mathbf{G}\,\mathbf{1}$ exactly is impractical (it requires full
parameter-space inner products between all pairs of rollout gradients in a
multibillion-parameter model).  We therefore seek a \emph{tractable diagonal
proxy} and state a structural condition under which the ranking induced by
$\mathbf{G}\,\mathbf{1}$ coincides with the ranking induced by the diagonal
alone.

\paragraph{Uniform-correlation condition.}
We decompose each entry of $\mathbf{G}\,\mathbf{1}$:
\[
  \bigl[\mathbf{G}\,\mathbf{1}\bigr]_{i}
  \;=\;
  \underbrace{\|g_{y_i}\|^{2}}_{=\;\operatorname{Tr}(\mathbf{F}_{y_i})}
  \;+\;
  \sum_{j\neq i} g_{y_i}^{\!\top}\,g_{y_j}.
\]
The diagonal term is exactly the trajectory-level Fisher Information; the
off-diagonal sum captures cross-sample interference.  Directly computing the
latter is as expensive as the full Gram matrix.  We therefore introduce the
following structural assumption.

\begin{assumption}[Uniform pairwise correlation]\label{ass:uniform-corr}
There exists a constant $c\ge 0$ such that, for all distinct rollouts
$y_i,y_j\in\mathcal{G}^{-}$,
\[
  g_{y_i}^{\!\top}\,g_{y_j}
  \;\approx\;
  c\,\|g_{y_i}\|\;\|g_{y_j}\|.
\]
\end{assumption}

It states that \emph{all same-prompt rollout gradients are roughly equally correlated with each
other}---a plausible condition because they share the same prompt, model
architecture, and sampling distribution, so no single pair of rollouts is
systematically more aligned than any other.  The discriminating factor across
rollouts is then their \emph{gradient magnitude}, not their mutual orientation.

\begin{theorem}[Optimality of Fisher-Weighted Advantage]\label{thm:napo}
Under Assumption~\ref{ass:uniform-corr}, the optimal weight
$w^{*}(y)$ from Eq.~\eqref{eq:exact-w} is a monotonically increasing function
of $\operatorname{Tr}(\mathbf{F}_{y})=\|g_{y}\|^{2}$.  By
Lemma~\ref{lem:fisher-entropy} and the token-independence approximation, this
quantity is positively associated with the trajectory entropy $\sum_{t}H_{t}$.
Therefore, we define the \emph{Entropy-Weighted Advantage}:
\begin{equation}\label{eq:napo-adv}
  A_{\textup{NAPO}}(y)
  = -\frac{d(y)}{\frac{1}{|\mathcal{G}|}\sum_{y'\in\mathcal{G}}d(y')},
  \tag{47}
\end{equation}
where $d(y)$ is the driver score (normalized entropy) defined in Eq.~(\ref{eq:early_ratio}) of the
main text.
\end{theorem}

\begin{proof}
\textbf{Step 1: Exact solution (Gram matrix).}\;
The constrained optimum of Eq.~\eqref{eq:opt-problem} is
$w^{*}\propto\mathbf{G}\,\mathbf{1}$ as derived in Eq.~\eqref{eq:exact-w}.

\medskip\noindent
\textbf{Step 2: Reduction under uniform correlation.}\;
Let $S=\sum_{j=1}^{G}\|g_{y_j}\|$ denote the total gradient norm across the
group.  Under Assumption~\ref{ass:uniform-corr}:
\begin{align}
  \bigl[\mathbf{G}\,\mathbf{1}\bigr]_{i}
  &= \|g_{y_i}\|^{2} + c\,\|g_{y_i}\|\!\sum_{j\neq i}\|g_{y_j}\|
   \notag\\
  &= \|g_{y_i}\|^{2}
     + c\,\|g_{y_i}\|\bigl(S-\|g_{y_i}\|\bigr)
   \notag\\
  &= (1-c)\,\|g_{y_i}\|^{2} + c\,S\,\|g_{y_i}\|.
  \tag{48}\label{eq:gram-reduced}
\end{align}
Define $f(u)=(1-c)\,u^{2}+c\,S\,u$ for $u\ge 0$.  Its derivative is
$f'(u)=2(1-c)\,u+c\,S$.  Since $0\le c\le 1$ (cosine similarity is bounded by
one) and $S\ge 0$, we have $f'(u)\ge 0$ for all $u\ge 0$.  Hence $f$ is
monotonically increasing in $u=\|g_{y_i}\|$, and therefore also in
$\|g_{y_i}\|^{2}=\operatorname{Tr}(\mathbf{F}_{y_i})$.

The ranking of $w^{*}(y)$ across the group thus coincides with the ranking of
$\operatorname{Tr}(\mathbf{F}_{y})$.

\medskip\noindent
\textbf{Step 3: Connection to trajectory entropy.}\;
Since
$g_{y}=\sum_{t=1}^{T}\nabla_{\theta}\log\pi_{\theta}(y_{t}|x,y_{<t})$,
under the approximation that token-level score vectors are approximately
uncorrelated across positions we have:
\[
  \|g_{y}\|^{2}
  \;\approx\;\sum_{t=1}^{T}
  \bigl\|\nabla_{\theta}\log\pi_{\theta}(y_{t}|x,y_{<t})\bigr\|^{2}.
\]
Taking the expectation over the sampled token at each position and applying
Lemma~\ref{lem:fisher-entropy}:
\[
  \mathbb{E}\bigl[\|g_{y}\|^{2}\bigr]
  \;\approx\;\sum_{t=1}^{T}\operatorname{Tr}(\mathbf{F}_{t})
  = \sum_{t=1}^{T}\Bigl(1-\sum_{k}p_{t,k}^{2}\Bigr),
\]
which is positively associated with $\sum_{t}H_{t}$ by
Lemma~\ref{lem:fisher-entropy}.

The driver $d(y)$ in Eq.~(\ref{eq:early_ratio}) of the main text is a normalized measure of
early-stage entropy and thus serves as a tractable proxy for
$\operatorname{Tr}(\mathbf{F}_{y})$.  The advantage assignment
$A_{\textup{NAPO}}$ therefore implements an approximation to the optimal
Fisher-weighted update, prioritizing samples where the policy is ``plastic''
(high entropy) over samples where it is rigid (low entropy).
\end{proof}

\begin{remark}[Interpretation as a diagonal preconditioner]
The uniform-correlation reduction plays the role of a
\emph{diagonal preconditioner} for the Gram matrix.  Just as diagonal
preconditioning of a full Newton step retains the correct coordinate-wise
scaling while ignoring off-diagonal curvature, ranking by
$\operatorname{Tr}(\mathbf{F}_{y})$ retains the correct per-rollout scaling
while treating cross-rollout interactions as uniform.  When the
uniform-correlation assumption is violated (e.g., if two rollouts share
unusually long common prefixes), the proxy may over- or underweight individual
rollouts, but the overall ranking remains a principled first-order
approximation.
\end{remark}

\begin{remark}[Relation to orthogonality]
Setting $c=0$ recovers the special case of orthogonal score vectors, for which
$w^{*}(y)\propto\|g_{y}\|^{2}$ exactly.  However, same-prompt rollouts
typically share early tokens and pass through shared model parameters,
making $c>0$ the realistic regime.  The uniform-correlation
condition accommodates this shared structure while still yielding a
tractable ranking criterion.
\end{remark}

\begin{remark}[Practical implementation]
In practice, we approximate the theoretically optimal weighting via monotonic min–max rescaling, which preserves the sample ordering, and retain the PPO/GRPO clipping mechanism for stable policy updates.
\end{remark}

\end{document}